\documentclass{article}

\PassOptionsToPackage{hyphens}{url}\usepackage{hyperref}

\usepackage[utf8]{inputenc} 
\usepackage[T1]{fontenc}    
\usepackage{booktabs}       
\usepackage{amsfonts}       
\usepackage{nicefrac}       
\usepackage{microtype}      
\usepackage{lipsum}
\usepackage{graphicx}
\usepackage{soul}
\usepackage{theapa}

\title{Transformer models: an introduction and catalog}

\author{
 Xavier Amatriain  \texttt{xavier@amatriain.net} \\
 Ananth Sankar \texttt{ansankar@linkedin.com} \\
  Jie Bing \texttt{jbing@linkedin.com} \\
  Praveen Kumar Bodigutla \texttt{pbodigutla@linkedin.com} \\
  Timothy J. Hazen  \texttt{thazen@linkedin.com} \\
  Michaeel Kazi \texttt{mkazi@linkedin.com} \\ \\
}

\begin{document}
\maketitle
\begin{abstract}
In the past few years we have seen the meteoric appearance of dozens 
of foundation models of the Transformer family, all of which have memorable and 
sometimes funny, but not self-explanatory, names. The goal of this paper is to offer a somewhat comprehensive but simple catalog and 
classification of the most popular Transformer models. The paper also 
includes an introduction to the most important aspects and innovations in Transformer models. 
Our catalog will include models that are trained using self-supervised learning (e.g., BERT or GPT3) as well as those that are further trained using a human-in-the-loop (e.g. the InstructGPT model used by ChatGPT).
\end{abstract}

\tableofcontents


\section{Introduction: What are Transformers}

Transformers are a class of deep learning models that are defined by some architectural traits. They were first introduced in the now famous "Attention is All you Need" paper (and associated blog 
post\footnote{\protect\url{https://ai.googleblog.com/2017/08/transformer-novel-neural-network.html}}) by Google researchers in 2017 \protect~\shortcite{vaswani2017attention}. The paper has accumulated a whopping 38k citations in only 5 years.

The original Transformer architecture is a specific instance of the encoder-decoder 
models~\shortcite{cho2014properties}\footnote{\protect\url{https://machinelearningmastery.com/encoder-decoder-long-short-term-memory-networks/}} that had become popular just over the 2–3 years prior. Up until that point however, attention was just one of the mechanisms used by these models, which were mostly based on LSTM (Long Short Term Memory)~\shortcite{hochreiter1997long} and other RNN (Recurrent Neural Networks)~\shortcite{mikolov2010recurrent} variations. The key insight of the Transformers paper, as the title implies, was that attention could be used as the only mechanism to derive dependencies between input and output.

The input to the Transformer is a sequence of tokens. The output of the encoder is a fixed-dimensional representation for each of the tokens along with a separate embedding for the sequence as a whole. The decoder takes the output of the encoder as input, and spits out a sequence of tokens as its output. In natural language processing (NLP), the tokens can be words or subwords. Subwords are used in all popular Transformer NLP models because they enable us to address the out-of-vocabulary (OOV) issue that is inherent in a word-based system. For simplicity, we will use the term "token" to refer to the items in the input and output sequences, understanding that these tokens are subwords for NLP systems. When Transformers are used for processing images or video, the tokens can represent sub-images or objects.

Since the publication of the paper, popular models like BERT and GPT have used only the encoder or decoder aspects of the original architecture. The core commonality of these models is, thus, not the encoder-decoder aspect, but, rather, the architecture of the individual layers in the encoders and decoders. The layer architecture of Transformers is based on a self-attention mechanism and a feed-forward layer, the core aspect of this being that each input token flows through the layers in its own path, while, at the same time, being directly dependent on every other token in the input sequence. This enables parallel and direct computation of contextual token representations which was previously not possible with sequential models like RNNs.

It is beyond the scope of this paper to go into all the details of the Transformer architecture. For that, we will refer you to the original paper \protect~\shortcite{vaswani2017attention} or to The Illustrated Transformer\footnote{\url{https://jalammar.github.io/illustrated-transformer/}} post. That being said, we will briefly describe the most important aspects since we will be referring to them in the catalog below. Let’s start with the basic architectural diagram from the original paper, and describe some of the components.

\subsection{Encoder/Decoder architecture}

\begin{figure}
    \centering
    \includegraphics[width=\textwidth,height=\textheight,keepaspectratio]{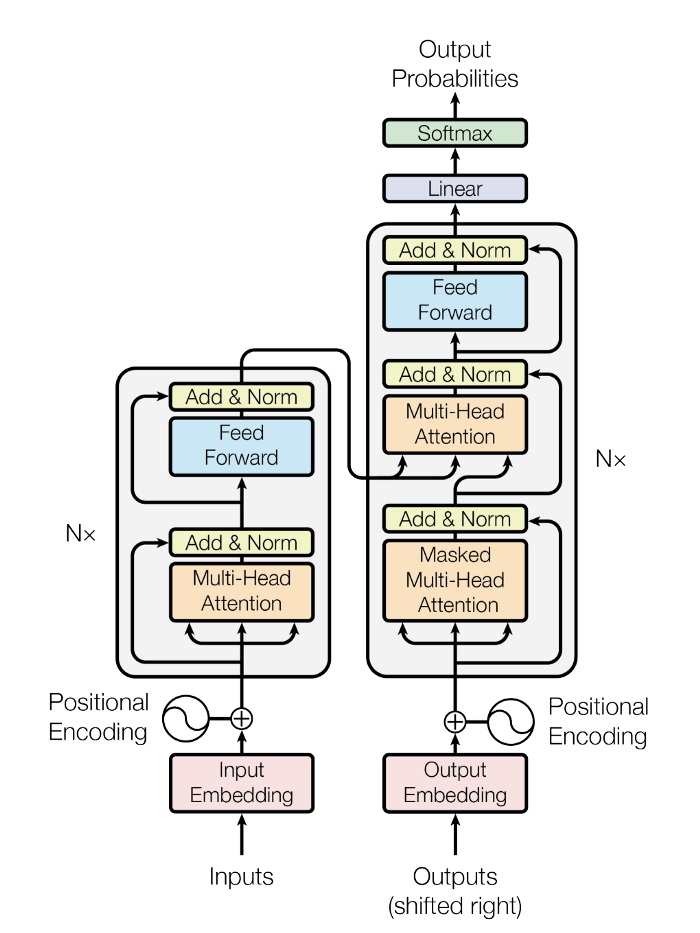}
    \caption{Transformer Architecture from {\protect\shortcite{vaswani2017attention}}}
    \label{fig:transformer}
\end{figure}

A generic encoder/decoder architecture (see Figure~\ref{fig:transformer}) is composed of two models. The encoder takes the input and encodes it into a fixed-length vector. The decoder takes that vector and decodes it into the output sequence. The encoder and decoder are jointly trained to 
maximize the conditional log-likelihood
of the output given the input. 
Once trained, the encoder/decoder can generate an output given an input sequence or can score a pair of input/output sequences.

In the case of the original Transformer architecture, both encoder and decoder had 6 identical layers. In each of those 6 layers the 
Encoder had two sub layers: a multi-head self attention layer, and a simple feed forward network. The self attention layer computes the output representation of each of its input tokens based on {\em all the input tokens}. Each sublayer also has a residual connection and a layer normalization. The output representation size of the Encoder was 512. The multi-head self-attention layer in the decoder is slightly different than that in the encoder. It masks all tokens to the right of the token whose representation is being computed so as to ensure that the decoder can only attend to tokens that come before the token it is trying to predict. This is shown in Figure~\ref{fig:transformer} as "masked multi-head attention." The Decoder also added a third sublayer, which is another multi-head attention layer over all the outputs of the Encoder. Note that all those specific details have since been modified in the many Transformer variations we will discuss. For example, as we noted before, models like BERT and GPT are based on only the encoder or decoder.

\subsection{Attention}

\begin{figure}
    \centering
    \includegraphics[width=\textwidth,height=\textheight,keepaspectratio]{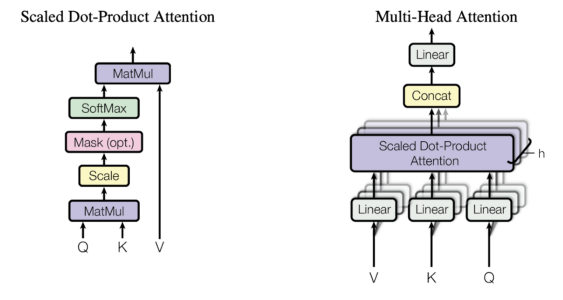}
    \caption{The Attention Mechanism from{\protect\shortcite{vaswani2017attention}}. (left) Scaled Dot-Product Attention, (right) Multi-Head Attention}
    \label{fig:attention}
\end{figure}

It is clear from the description above that the only “exotic” elements of the model architecture are the 
multi-head attention layers, 
but, as described above, that is where the whole power of the model lies! So, what is attention anyway? An attention function is a mapping between a query and a set of key-value pairs to an output. 
Each token in the input to the attention layer is converted to a query, key and value using three corresponding matrices. The output representation of each token is computed as a weighted sum of the values of all the tokens, where the weight assigned to each value is computed by a compatibility function of its associated key and the query of the token whose representation is being computed. The compatibility function used in Transformers is just a scaled dot product. A key aspect of this attention mechanism in Transformers is that each token flows through its own computation path, thus lending itself to parallel computation of the representation of all the tokens in the input sequence. Now that we understand how attention works, what is multi-head attention? Well, that is just multiple attention blocks independently computing representations for each token. All these representations are then aggregated to give the final representation of the token. We will refer you again to the The Illustrated Transformer\footnote{\url{https://jalammar.github.io/illustrated-transformer/}} post for many more details on how the attention mechanism works, but will reproduce the diagram from the original paper in Figure~\ref{fig:attention} so you get the main idea.

There are several advantages of attention layers over recurrent and convolutional networks, the two most important being their lower computational complexity and their higher connectivity, especially useful for learning long-term dependencies in sequences.

\subsection{Foundation vs Fine-tuned models}

A foundation model is defined as "any model that is trained on broad data (generally using self-supervision at scale) that can be adapted (e.g., fine-tuned) to a wide range of downstream tasks”~\shortcite{bommasani2021opportunities}. When the foundation model is further trained on a small amount of target-specific data, it is called a fine-tuned model\footnote{\url{https://huggingface.co/docs/transformers/training}} because it has been fine-tuned to the specifics of the task at hand.

The BERT paper~\shortcite{devlin2018bert} popularized this approach of pretraining and finetuning for natural language processing, resulting in many researchers using this approach for many different tasks. As a consequence, most of the leaderboards for any language-related machine leartning (ML) task became completely dominated by some version of the Transformer architecture (see for example the well known SQUAD leaderboard\footnote{\url{https://rajpurkar.github.io/SQuAD-explorer}} for question answering or the GLUE leaderboard\footnote{\url{https://gluebenchmark.com/leaderboard}} for general language understanding, where all systems at the top employ Transformer-based models).

In its original usage, "fine-tuning" referred to tweaking a foundation model for a specific task, such as spam classification or question answering. Models, such as BERT, produce representations of the input tokens, but do not, by themselves, accomplish any task. Thus, it is necessary to fine-tune them by adding extra neural layers on top of the foundation model and training the model end to end. 

With generative models like GPT, things are a little different. GPT is a decoder language model trained to predict the next token of a sentence given all the previous tokens. By training on huge amounts of web corpora covering almost any topic one can think about, it was found that GPT could actually produce reasonable outputs to input queries or prompts. GPT accomplished this by simply predicting the next token given the input prompt sequence and the output sequence GPT had already predicted. This language generation actually did a somewhat reasonable job of tasks like answering questions about general web knowledge, writing poems etc. Notwithstanding, GPT's outputs were often untruthful or really not very helpful to the user. To address this, OpenAI researchers came up with the idea of training GPT to follow human instructions~\shortcite{ouyang2022training}. The resulting models are called InstructGPT. The authors did this by using a small amount of human-labeled data from a large variety of tasks to further train GPT. As before, this is a "fine-tuning" process, but the resulting Instruct GPT model is capable of doing a wide range of tasks, and is, in fact, the class of models used by the popular ChatGPT engine. Since these models can accomplish a myriad of tasks, we refer to them as foundation models.

Such additional fine-tuning has been used to generate other general purpose model variants as well, specifically designed for uses cases beyond language modeling (predicting the next token in a sequence). For example, there is a subclass of models fined-tuned to learn text string embeddings optimized for semantic-relatedness, making them directly useful for higher-level semantic tasks (e.g. text classification, clustering, search retrieval, etc.). Examples include OpenAI's text embedding models\footnote{\url{https://platform.openai.com/docs/guides/embeddings/what-are-embeddings}}, E5\footnote{\url{https://huggingface.co/intfloat/e5-large}}, and InstructOR\footnote{\url{https://huggingface.co/hkunlp/instructor-xl}}. Transformer encoders have also been successfully fined-tuned within multi-task learning frameworks to be able to perform multiple different semantic tasks using a single shared Transformer model ~\shortcite{liu2019mtdnn,aghajanyan2021muppet}.

Thus, as we see, while originally foundation models were fine-tuned for very specific target tasks for specific groups of users, today fine-tuning is used to also create further versions of foundation models that can be used by a huge number of users. The process used by ChatGPT and similar dialog agents, like BlenderBot3 or Sparrow, is fairly simple: Given a pretrained language model like GPT, we use it to generate different responses to input prompts (or instructions) and have humans rank the results. We then use those rankings (aka preferences or feedback) to train a reward model. The reward model attaches a score to each output for a given input instruction. After this, a reinforcement learning with human feedback (RLHF) process~\shortcite{christiano2023deep} is used to train the model on more input instructions, but, rather than use a human to generate the feedback, the reward model is used to rank the outputs of the model. You can read much more in these two wonderful posts by Huggingface\footnote{\url{https://huggingface.co/blog/rlhf}} and Ayush Thakur\footnote{\url{https://wandb.ai/ayush-thakur/RLHF/reports/Understanding-Reinforcement-Learning-from-Human-Feedback-RLHF-Part-1--VmlldzoyODk5MTIx}}.

\begin{figure}
    \centering
    \includegraphics[width=\textwidth,height=\textheight,keepaspectratio]{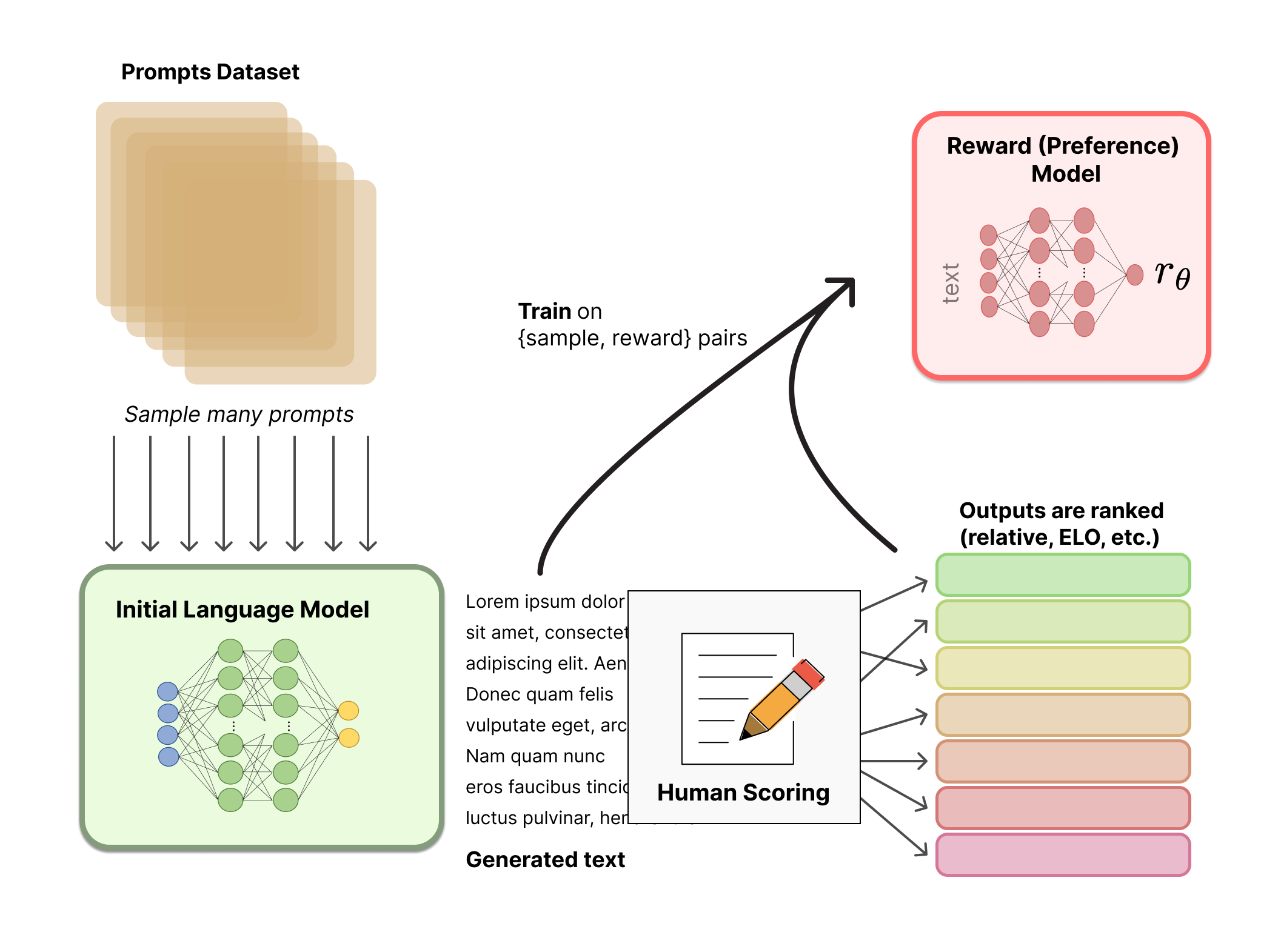}
    \caption{Reinforcement Learning with Human Feedback. From HuggingFace’s RLHF blog post at \protect\url{https://huggingface.co/blog/rlhf}}
    \label{fig:rlhf}
\end{figure}

\subsection{The Impact of Transformers}

The application demonstrated in the original Transformer paper~\shortcite{vaswani2017attention} was language translation. This seminal work also showed the architecture generalized well to other language tasks. Over the next several months, researchers figured out that Transformers could be used to capture a lot of inherent knowledge about language by pretraining them on a very large amount of unsupervised text. The knowledge captured in these models could then be transferred to target tasks by training on a small amount of labeled data. 

While original Transformers were designed for language tasks, the same Transformer architecture has been applied to many other applications like the generation of images, audio, music, or even actions. Because of that, Transformers are considered a key, if not the key, component to the new wave of the so-called "Generative AI". Generative AI and its many applications are already revolutionizing many aspects of society~\shortcite{stokelwalker2023nature,baidoo2023education}

Of course all these applications would not have been possible but for the myriad of tools that made them readily available to anyone that could write a few lines of code. Not only were Transformers quickly integrated into the main AI frameworks (namely Pytorch\footnote{\url{https://pytorch.org/tutorials/beginner/transformer_tutorial.html}} and TensorFlow (TF)\footnote{\url{https://www.tensorflow.org/text/tutorials/transformer}}), but they even enabled the creation of an entire company around them. Huggingface\footnote{\url{https://huggingface.co/docs}}, a startup that has raised over \$ 60M to this day, is almost entirely built around the idea of commercializing their open source Transformers library\footnote{\url{https://github.com/huggingface/transformers}}. 

Transformer model adoption is further accelerated as specialized hardware is developed by commercial players to improve model training and inference speed. NVIDIA's Hopper Tensor Cores\footnote{\url{https://resources.nvidia.com/en-us-tensor-core/nvidia-tensor-core-gpu-datasheet}} can apply mixed FP8 and FP16 precisions to dramatically accelerate AI calculations for Transformers.

Last but not least, we would be remiss if we did not mention the impact of ChatGPT on the popularization of Transformers and Large Langage Models (LLMs) in particular\cite{minaee2024large}. ChatGPT was released by OpenAI in November 2022, and became the fastest growing app in history, reaching 1 million users in less than a month, and 100 million in less than two~\shortcite{UBSChatGPT2023}. ChatGPT was originally a chatbot application built on top of the Instruct-GPT model~\shortcite{ouyang2022training} also called GPT-3.5. Not much later, OpenAI announced the release of the more powerful GPT-4\footnote{https://openai.com/research/gpt-4}, which achieves human capabilities in tasks such as passing the USMLE exam for medical doctors or the bar exam for lawyers~\shortcite{gpt-4}.

\subsection{A Note on Diffusion models}

Diffusion models have become the new state-of-the-art in image generation, clearly pushing aside the previous approaches such as GANs (Generative Adversarial Networks). It is important to note, though, that the diffusion mechanism is not dependent on the Transformer architecture. However, most modern diffusion approaches do include a Transformer backbone~\shortcite{esser2021taming}.

\begin{figure}
    \centering
    \includegraphics[width=\textwidth,height=\textheight,keepaspectratio]{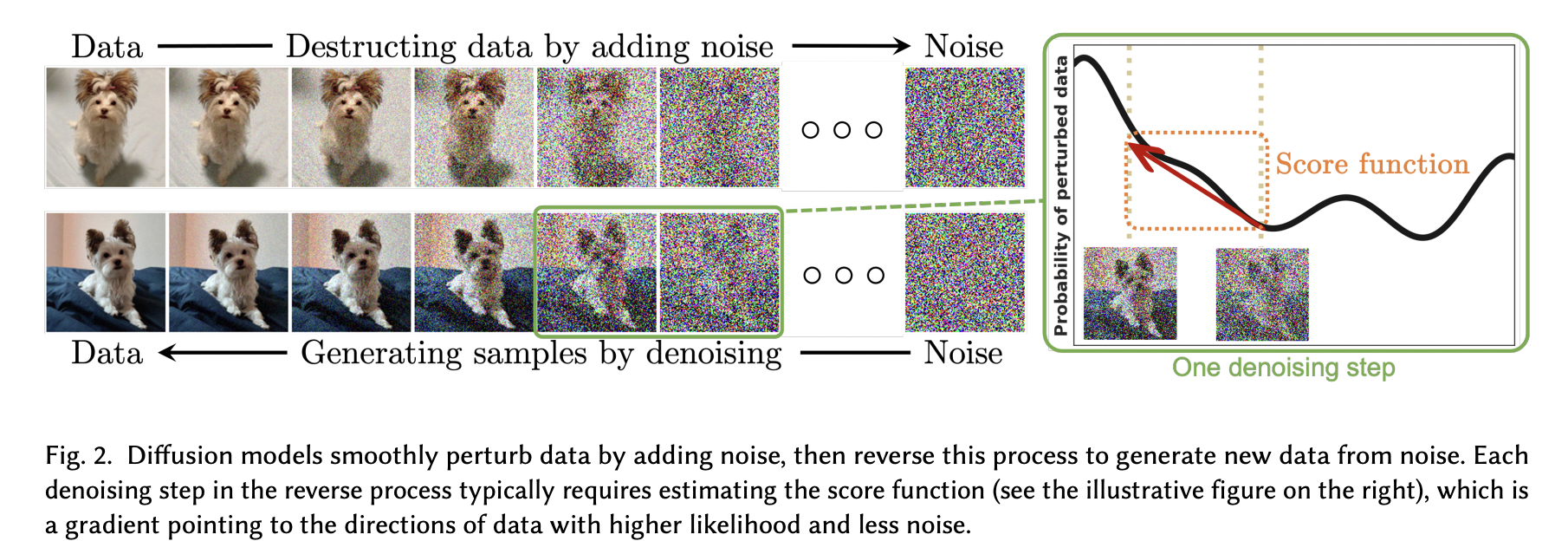}
    \caption{Probabilistic diffusion model architecture from “Diffusion Models: A Comprehensive Survey of Methods and Applications," Figure 2 \protect\shortcite{yang2022diffusion} }
    \label{fig:diffusion}
\end{figure}

Diffusion models are a class of latent variable models trained through variational inference. What this means in practice is that we train a deep neural network to denoise images blurred with some sort of noise function. Networks that are trained this way are in fact learning the latent space of what those images represent (see Figure~\ref{fig:diffusion}).

Diffusion models have relation to other generative models like Denoising Autoencoders and the famous Generative Adversarial Networks (GAN)\footnote{\url{https://en.wikipedia.org/wiki/Generative_adversarial_network}}, which they have mostly replaced in many applications. Some authors\footnote{\url{https://benanne.github.io/2022/01/31/diffusion.html}} will go as far as saying that Diffusion models are just a specific instance of autoencoders. However, they also admit that the small differences do transform their application, from the latent representation of autoencoders to the pure generative nature of Diffusion models.

\section{The Transformers Catalog}

\subsection{Features of a Transformer}

In this section we will introduce a catalog of the most important Transformer models that have been developed to this day. We will categorize each model according to the following properties: Family, Pretraining Architecture, Pretraining or Fine-tuning Task, Extension, Application, Date (of first known publication), Number of Parameters, Corpus, License, and Lab. Some are relative simple to understand: \emph{Family} represents what original foundation model the specific model is extending, \emph{extension} describes what the model is adding to the one it is deriving from, \emph{Date} is when the model was firts published, \emph{Number of parameters} of the pretrained model, \emph{Corpus} is what data sources the model was pre-trained or fine-tuned on, \emph{License} describes how the model can be legally used, and \emph{Lab} lists the institution that published the model. The remaining propterties deserve a bit more explanation. We do that in the paregraphs that follow: 

\subsubsection{Pretraining Architecture}

We described the Transformer architecture as being made up of an Encoder and a Decoder, and that is true for the original Transformer. However, since then, different advances have been made that have revealed that in some cases it is beneficial to use only the encoder, only the decoder, or both.

\paragraph{Encoder Pretraining}

These models, which are also called bi-directional or auto-encoding, only use the encoder during pretraining, which is usually accomplished by masking tokens in the input sentence and training the model to 
reconstruct those tokens. 
At each stage during pretraining, self-attention layers can access all their input tokens. This family of models are most useful for tasks that require understanding 
complete sentences or passages, such as text classification, entailment, and extractive question answering.

\paragraph{Decoder Pretraining}

Decoder models use only the decoder during a pretraining. They are also called auto-regressive language models because they are trained to predict the next token based on the previous sequence of tokens. 

The self-attention layers can only access the tokens positioned before a given token in the sentence. They are best suited for tasks involving text generation.

\paragraph{Transformer (Encoder-Decoder) Pretraining}

Encoder-decoder models, also called sequence-to-sequence, use both parts of the Transformer architecture. 

Self-attention layers of the encoder can access all their input tokens, while the self-attention layers of the decoder can only access the tokens positioned before a given token. As explained before, the additional attention layer in the decoder enables access to all encoder token representations. 

An encoder-decoder model can be pre-trained by optimizing denoising objectives~\shortcite{lewis2019bart} or a combination of denoising and causal language modeling objectives~\shortcite{soltan2022alexatm}. These objective functions are complex in comparison to the ones used to pretrain encoder only or decoder only models. 
The encoder-decoder models are best suited for tasks revolving around generating new sentences depending on a given input, such as summarization, translation, or generative question answering.

\subsubsection{Pretraining  or Finetuning Task}

When training a model we need to define an objective, or task, for the model to learn on. Some of the typical tasks, such as predicting the next token or learning to reconstruct masked tokens were already mentioned above. “Pre-trained Models for Natural Language Processing: A Survey”~\shortcite{qiu2020pre} includes a pretty comprehensive taxonomy of pretraining tasks, all of which can be considered self-supervised:

\begin{enumerate}
    \item \textbf{Language Modeling (LM):} Predict the next token (in the case of unidirectional LM) or the previous and next token (in the case of bidirectional LM).
    \item \textbf{Causal Language Modeling (Causality-masked LM):} Autoregressively (left-to-right, in general) predict a text sequence, similar to unidirectional LM.
    \item \textbf{Prefix Language Modeling (Prefix LM):} In this task, a separate 'prefix' section is separated from the main sequence. Within the prefix, any token can attend to any other token (non-causal). Outside of the prefix, decoding proceeds autoregressively.
    \item \textbf{Masked Language Modeling (MLM):} Mask out some tokens from the input sentences and then train the model to predict the masked tokens using the surrounding context.
    \item \textbf{Permuted Language Modeling (PLM):} Same as LM, but on a random permutation of input sequences. A permutation is randomly sampled from all possible permutations. Then some of the tokens are chosen as the target, and the model is trained to predict these targets.
    \item \textbf{Denoising Autoencoder (DAE):} Take a partially corrupted input and aim to recover the original, undistorted input. Examples of corrupted input include randomly sampling tokens from the input and replacing them with "[MASK]" elements, randomly deleting tokens from the input, or shuffling sentences in random order.
       
       \item \textbf{Replaced Token Detection (RTD):} Using a "generator" model, randomly replace certain tokens in the text. The "discriminator" is tasked to predict whether a token comes from the original text, or the generator model.
       \item \textbf{Next Sentence Prediction (NSP):} Train the model to distinguish whether two input sentences are continuous segments from the training corpus.
       
   
\end{enumerate}

Note that in the case of fine-tuned models, this property is used to describe the task the model was fine-tuned to, not how it was pre-trained.

\subsubsection{Application}

Here we will note what are the main practical applications of the Transformer model. Most of these applications will be in the language domain (e.g. question answering, sentiment analysis, or entity recognition). However, as mentioned before, some Transformer models have also found applications well beyond NLP and are also included in the catalog.

\subsection{Catalog table}

You can access a table format of the catalog at \protect\url{http://bit.ly/3YFqRn9} for easier browsing across the different model features.

\subsection{Family Tree}

\begin{figure}
    \centering
    \includegraphics[width=\textwidth,height=\textheight,keepaspectratio]{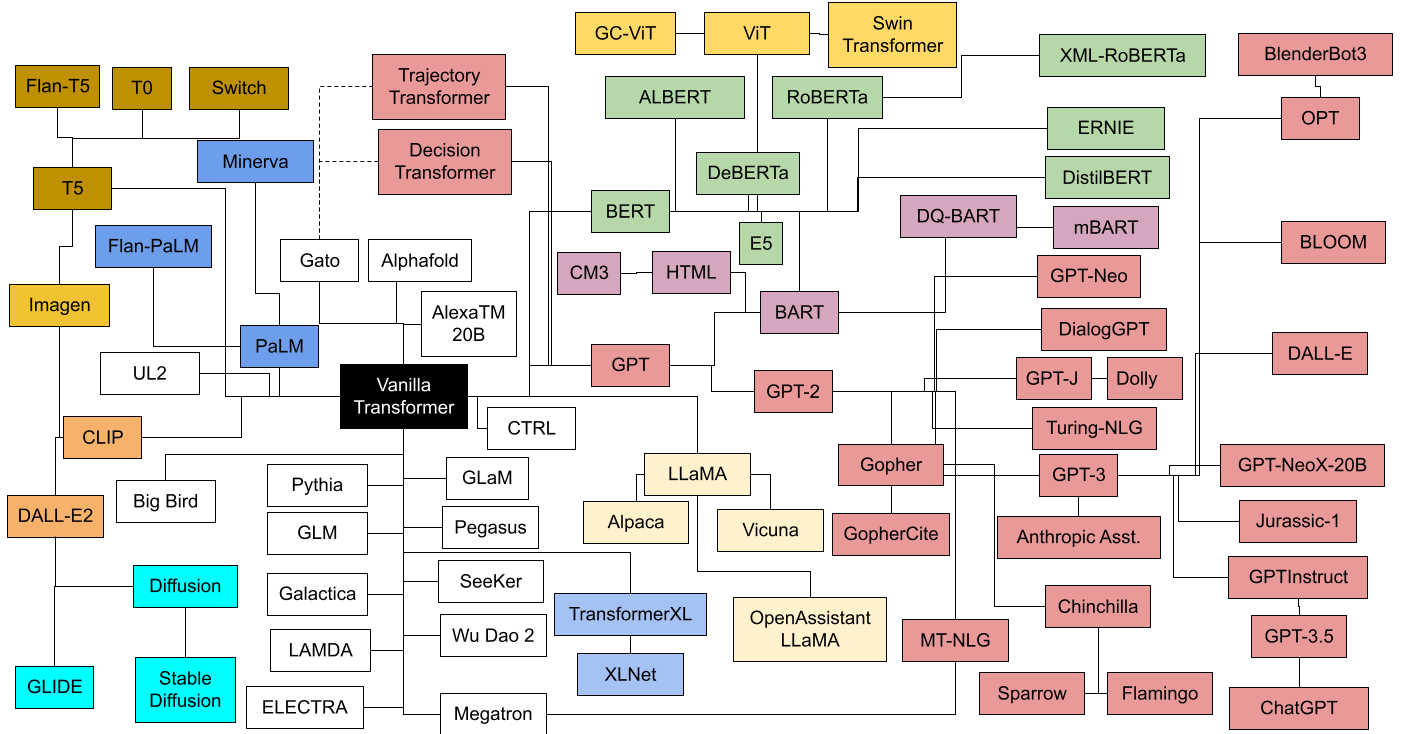}
    \caption{Transformers Family Tree}
    \label{fig:tree}
\end{figure}

The diagram in Figure~\ref{fig:tree} is a simple view that highlights the different families of Transformers and how they relate to each other.

\subsection{Chronological timeline}

\begin{figure}
    \centering
    \includegraphics[width=\textwidth,height=\textheight,keepaspectratio]{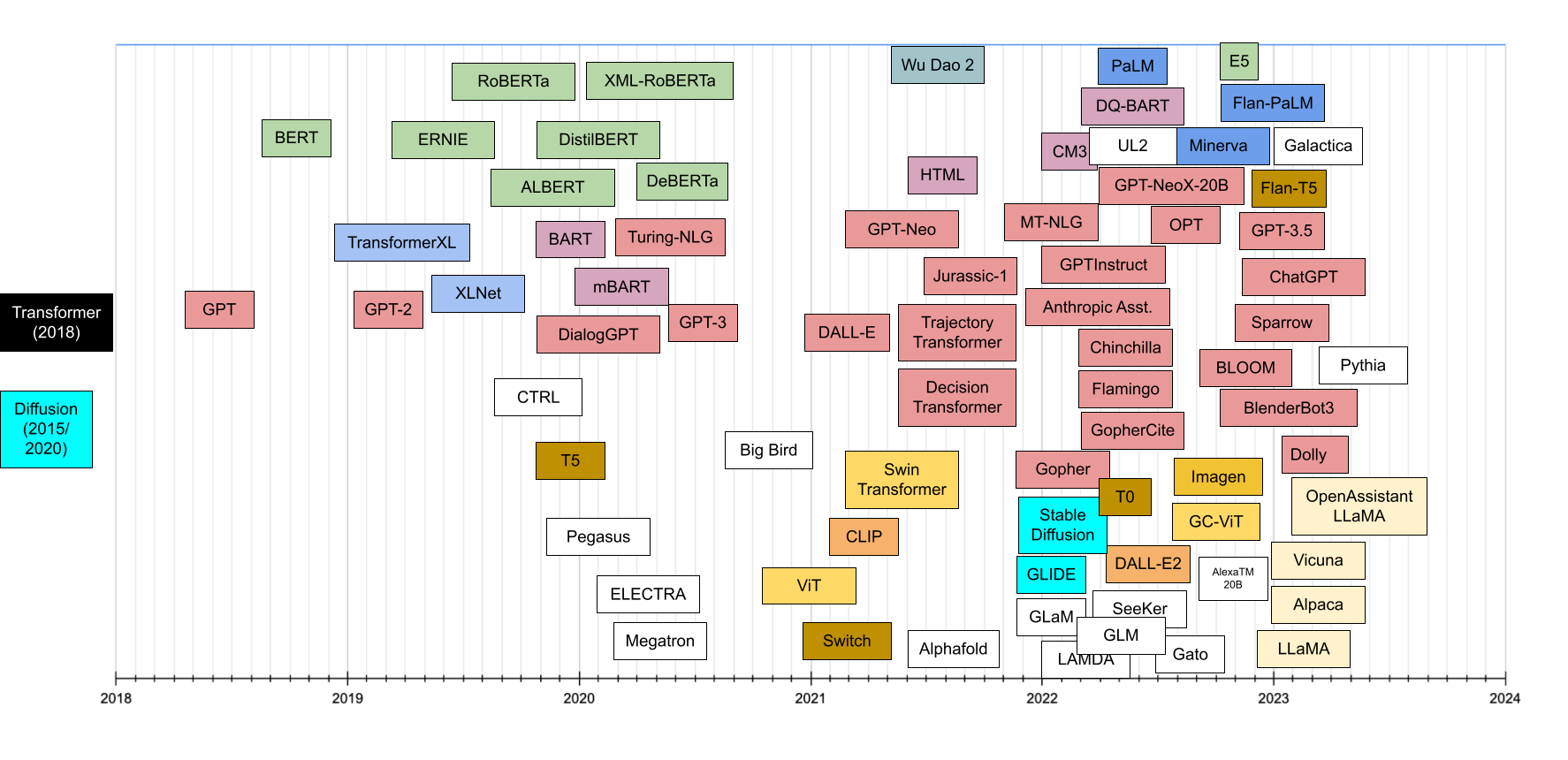}
    \caption{Transformer timeline. Colors describe Transformer family.}
    \label{fig:timeline}
\end{figure}

Another interesting perspective of the catalog is to see it as a chronological timeline. In Figure~\ref{fig:timeline} you will find all the Transformers in the catalog sorted by their date of publication. In this first visualization, the Y-axis is only used to cluster Transformers of related heritage/family.

\begin{figure}
    \centering
    \includegraphics[width=\textwidth,height=\textheight,keepaspectratio]{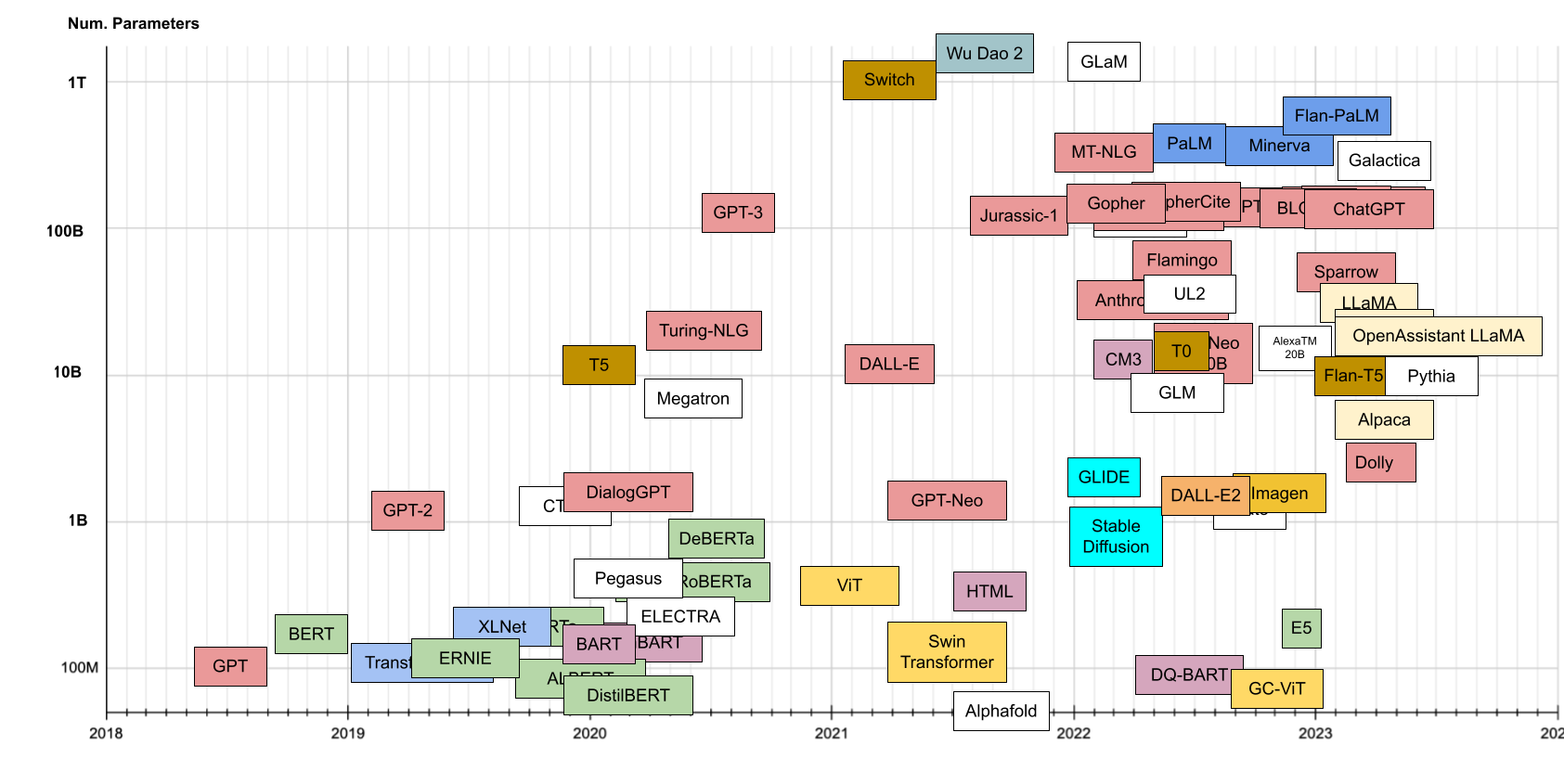}
    \caption{Transformer timeline. On the vertical axis, number of parameters. Colors describe Transformer family.}
    \label{fig:timelineSize}
\end{figure}

In Figure~\ref{fig:timelineSize}, the Y-axis represents model size in millions of parameters. You won't be able to see all the models in the catalog since many fall right on the same time and size, so please refer to the previous image for that.

\begin{figure}
    \centering
    \includegraphics[width=\textwidth,height=\textheight,keepaspectratio]{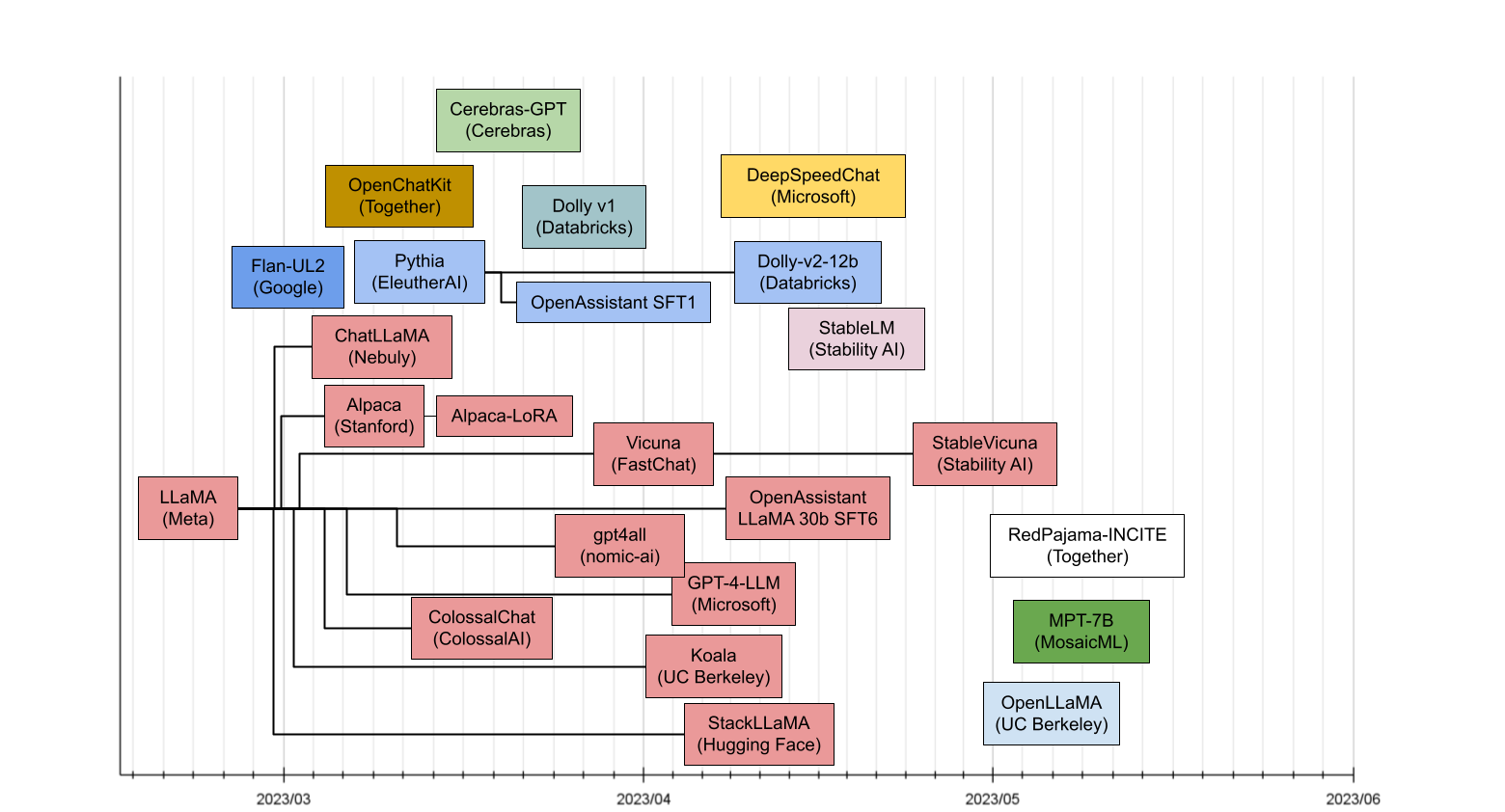}
    \caption{Recently published LLMs}
    \label{fig:finetunedModels}
\end{figure}

Since the introduction of chatGPT, the LLM open-source community has experienced a significant surge in activity. With each passing week, we have observed a proliferation of refined models fine-tuned using the latest technologies. As a result, these models are continuously improving, growing more robust and powerful. Figure~\ref{fig:finetunedModels} demonstrates the recent emerged models since Feb, 2023.

\appendix
\section{Catalog List}

Finally, here is the full list view that might be easier to follow along in some cases:

\subsection{ALBERT}
            \begin{itemize}             
                \item \textbf{Reference:} \href{https://arxiv.org/abs/1909.11942}{\shortcite{lan2019albert}}
                \item \textbf{Link:} \url{https://huggingface.co/docs/transformers/model_doc/albert}
                \item \textbf{Family:} BERT
                \item \textbf{Pretraining Architecture:} Encoder
                \item \textbf{Pretraining Task:} MLM/NSP
                \item \textbf{Extension:} Compressed version of BERT using parameter sharing, which is much more efficient given the same number of parameters
                \item \textbf{Application:} Same as BERT
                \item \textbf{Date (of first known publication):} 09/2019
                \item \textbf{Num. Params:} Base = 12M, Large = 18M, XLarge = 60M*
                \item \textbf{Corpus:} Same as BERT
                \item \textbf{License:} Open, Apache-2.0
                \item \textbf{Lab:} Google
            \end{itemize}

\subsection{AlexaTM 20B}
           \begin{itemize}
            \item \textbf{Reference:} \shortcite{soltan2022alexatm}
            \item\textbf{Link:} \url{https://github.com/amazon-science/alexa-teacher-models}
            \item \textbf{Family:} Transformer
            \item \textbf{Pretraining Architecture:} Encoder/Decoder
            \item \textbf{Pretraining Task:} Optimizes denoising ($80\%$) and Prefix LM ($20\%$)
            \item \textbf{Extension:} Derived from BART and layernorms located exactly at the beginning of each layer. Encoder initialized with internal 10B pre-trained encoder.
            \item \textbf{Application:} Summarization, multi-lingual machine translation and NLU tasks
            \item \textbf{Date (of first known publication):} 08/2022
            \item \textbf{Num. Params:} 20B
            \item \textbf{Corpus:}   Wikipedia and mC4 datasets in 12 languages.
            \item \textbf{License:} Limited, non-commercial
            \item \textbf{Lab:} Amazon
            \end{itemize}

\subsection{Alpaca}
            \begin{itemize}
            \item \textbf{Reference:}~\shortcite{alpaca}
            \item \textbf{Link:} \url{https://github.com/tatsu-lab/stanford_alpaca}
            \item \textbf{Family:} LLaMA
            \item \textbf{Pretraining Architecture:} Decoder
            \item \textbf{Fine-tuning Task:} human instructions 
            \item \textbf{Extension:} Alpaca is fine-tuned from a 7B LLaMA model.
            \item \textbf{Application:} Evaluated on a variety of text generation and classification tasks.
            \item \textbf{Date (of first known publication):} 03/2023
            \item \textbf{Num. Params:} 7B
            \item \textbf{Corpus:} 52K instruction-following data generated using self-instruct mechanism, from 175 human-written instruction-output pairs. 
            \item \textbf{License:} Limited, Non-commercial bespoke license
            \item \textbf{Lab:} Stanford
            \end{itemize}
            
\subsection{AlphaFold}
            \begin{itemize}
            \item \textbf{Reference:}~\shortcite{jumper2021highly}          
            \item \textbf{Link:}~\url{https://github.com/deepmind/alphafold}
            \item \textbf{Family:} SE(3) Transformer~\shortcite{fuchs2020se3}
            \item \textbf{Pretraining Architecture:} Encoder
            \item \textbf{Pretraining Task:} Protein folding prediction of BERT using parameter sharing, which is much more efficient given the same number of parameters
            \item \textbf{Extension:} The original Alphafold used a BERT-style Transformer. The details of Alphafold’s Transformer are not known, but it is believed it is an extension of the SE(3)-Tranformer, a 3-D equivariant Transformer (see this blog post\footnote{\url{https://fabianfuchsml.github.io/alphafold2/}})   
            \item \textbf{Application:} Protein folding
            \item \textbf{Date (of first known publication):} 09/2019
            \item \textbf{Num. Params:}b12M, Large = 18M, XLarge = 60M*
            \item \textbf{Corpus:} Same as BERT
            \item \textbf{License:} the code is open sourced, with Apache-2.0
            \item \textbf{Lab:} Deepmind
            \end{itemize}

\subsection{Anthropic Assistant}
            \begin{itemize}
                \item \textbf{Reference:}~\shortcite{bai2022training,askell2021general,bai2022constitutional}
                \item \textbf{Link:} N/A
                \item \textbf{Family:} Transformer 
                \item \textbf{Pretraining Architecture:} Decoder
                \item \textbf{Pretraining Task:} LM
                \item \textbf{Extension:} These models do not introduce novelties at the architecture/pretraining level and they are similar to GPT-3, but they focus on how to improve alignment through fine-tuning and prompting. Note that the Anthropic Assistant includes several models optimized for different tasks. The work often focus on the benefits of RLHF. Latest versions of this work study using an LLM to critique the model output for harmlessness, and provide feedback data for RL this way (RLHF -> RLAIF).   
                \item \textbf{Application:} Different models with different applications from general dialog to code assistant.
                \item \textbf{Date (of first known publication):} 12/2021
                \item \textbf{Num. Params:}10M to 52B
                \item \textbf{Corpus:} 400B tokens from filtered Common Crawl and Books, and 10\% python code. They also create several Dialogue Preference datasets for the RLHF training.
                \item \textbf{License:} N/A
                \item \textbf{Lab:} Anthropic
            \end{itemize}

\subsection{BART}
            \begin{itemize}
                \item \textbf{Reference:} \shortcite{lewis2019bart}
                \item \textbf{Link:} \url{https://huggingface.co/docs/transformers/model_doc/bart}
                \item \textbf{Family:} BERT for encoder, GPT for Decoder 
                \item \textbf{Pretraining Architecture:} Encoder/Decoder
                \item \textbf{Pretraining Task:} DAE
                \item \textbf{Extension:} It can be seen as a generalization of BERT and GPT in that it combines ideas from both in the encoder and decoder   
                \item \textbf{Application:} Mostly text generation but also some text understanding tasks*
                \item \textbf{Date (of first known publication):} 10/2019*
                \item \textbf{Num. Params:} Base = 140M, Large = 400M. In general, roughly 10\% larger than BART for equivalent architectures.
                \item \textbf{Corpus:}Same as RoBERTa (160Gb of news, books, stories)
                \item \textbf{License:} Open, Apache-2.0
                \item \textbf{Lab:}Facebook
            \end{itemize}
            
\subsection{BERT}

            \begin{itemize}
                \item \textbf{Reference:} \shortcite{devlin2018bert}
                \item \textbf{Link:} \url{https://huggingface.co/docs/transformers/model_doc/bert}
                \item \textbf{Family:} BERT 
                \item \textbf{Pretraining Architecture:} Encoder
                \item \textbf{Pretraining Task:} MLM/NSP
                \item \textbf{Extension:It can be seen as a generalization of BERT and GPT in that it combines ideas from both in the encoder and decoder}   
                \item \textbf{Application:}General Language Understanding and Question Answering. Many other language applications followed
                \item \textbf{Date (of first known publication):} 10/2018
                \item \textbf{Num. Params:}Base = 110M, Large = 340MT
                \item \textbf{Corpus:}Toronto Book Corpus and Wikipedia (3.3B Tokens)
                \item \textbf{License:} Open, Apache-2.0
                \item \textbf{Lab:}Google
            \end{itemize}

\subsection{Big Bird}

            \begin{itemize}
                \item \textbf{Reference:} \shortcite{zaheer2020big}
                \item \textbf{Link:} \url{https://huggingface.co/docs/transformers/model_doc/big_bird}
                \item \textbf{Family:} BERT 
                \item \textbf{Pretraining Architecture:} Encoder
                \item \textbf{Pretraining Task:} MLM
                \item \textbf{Extension:} Big Bird can extend other architectures such as BERT, Pegasus, or RoBERTa by using a sparse attention mechanism that elminates the quadratic dependency thus making it more suitable for longer sequences 
                \item \textbf{Application:}Particularly well suited for longer sequences, not only in text but also e.g. in genomics
                \item \textbf{Date (of first known publication):} 07/2020
                \item \textbf{Num. Params:}Depends on the overall architecture
                \item \textbf{Corpus:}Books, CC-News, Stories and Wikipedia)
                \item \textbf{License:} Open, Apache-2.0
                \item \textbf{Lab:}Google
            \end{itemize}

\subsection{BlenderBot3}

            \begin{itemize}
                \item \textbf{Reference:} \shortcite{shuster2022blenderbot}
                \item \textbf{Link:} \url{https://parl.ai/projects/bb3/}
                \item \textbf{Family:} GPT 
                \item \textbf{Pretraining Architecture:} Decoder
                \item \textbf{Pretraining Task:} LM
                \item \textbf{Extension:} BlenderBot 3 is based on a pre-trained OPT. It adds features needed for a dialog agent such as long-term memory or the ability to search the internet. It is also fine-tuned for some specific tasks given human feedback on them.  
                \item \textbf{Application:} Same as GPT-3
                \item \textbf{Date (of first known publication):} 08/2022
                \item \textbf{Num. Params:} 3B, 30B, and 175B
                \item \textbf{Corpus:} 180B tokens = RoBERTa + the Pile + PushShift.io Reddit
                item \textbf{License:} Limited, non-commercial, research only
                \item \textbf{Lab:}Facebook
            \end{itemize}

\subsection{BLOOM}
            \begin{itemize}
                \item \textbf{Reference:} See blog post\footnote{\url{https://huggingface.co/blog/bloom-inference-optimization}}
                \item \textbf{Link:} \url{https://huggingface.co/docs/transformers/model_doc/bloom}
                \item \textbf{Family:} GPT 
                \item \textbf{Pretraining Architecture:} Decoder
                \item \textbf{Pretraining Task:} LM
                \item \textbf{Extension:} Main difference to GPT-3 is that it uses full attention instead of sparse attention  
                \item \textbf{Application:} Same as GPT-3
                \item \textbf{Date (of first known publication):} 07/2022
                \item \textbf{Num. Params:}176B
                \item \textbf{Corpus:} 366B tokens (1.5 TB of text data) multilingual dataset
                \item \textbf{Lab:} Big Science/Huggingface
                \item \textbf{License:} Open, but need to follow restrictions in Attachment A, BigScience RAIL License v1.0
            \end{itemize}
            
\subsection{ChatGPT}

            \begin{itemize}
                \item \textbf{Reference:} See blog post\footnote{\url{https://openai.com/blog/chatgpt/}}
                \item \textbf{Link:} \url{https://chat.openai.com}
                \item \textbf{Family:} GPT 
                \item \textbf{Pretraining Architecture:} Decoder
                \item \textbf{Pretraining Task:} LM
                \item \textbf{Extension:} ChatGPT takes a GPT3.5 (aka GPT3 Davinci-003) pretrained model and uses RLHF to finetune the model mostly like described in InstructGPT but with slight differences in the data collection. ChatGPT is also more than a model since it includes extensions for Memory Store and retrieval similar to BlenderBot3  
                \item \textbf{Application:} Dialog agents
                \item \textbf{Date (of first known publication):} 10/2022
                \item \textbf{Num. Params:} Same as GPT3
                \item \textbf{Corpus:} Same as GPT3 + datasets generated for RLHF
                \item \textbf{License:} Closed source, accessible through API
                \item \textbf{Lab:} OpenAI
            \end{itemize}

\subsection{Chinchilla}

            \begin{itemize}
                \item \textbf{Reference:} \href{https://arxiv.org/abs/2203.15556}{\shortcite{hoffmann2022training}}
                \item \textbf{Link:} N/A
                \item \textbf{Family:} GPT 
                \item \textbf{Pretraining Architecture:} Decoder
                \item \textbf{Pretraining Task:} LM
                \item \textbf{Extension:} Same as Gopher but with optimizations to reduce model size and therefore training/inference time with equal or superior performance  
                \item \textbf{Application:} Same as Gopher/GPT3
                \item \textbf{Date (of first known publication):} 03/2022
                \item \textbf{Num. Params:}70B
                \item \textbf{Corpus:} Massive Text
                \item \textbf{License:} Closed source.
                \item \textbf{Lab:} Deepmind
            \end{itemize}

\subsection{CLIP}

            \begin{itemize}
                \item \textbf{Reference:} \shortcite{radford2021learning}
                \item \textbf{Link:} \url{https://huggingface.co/docs/transformers/model_doc/clip}
                \item \textbf{Family:} CLIP (Also using Resnet, ViT, and vanilla Transformer for text) 
                \item \textbf{Pretraining Architecture:} Encoder
                \item \textbf{Pretraining Task:} predict which of the N × N possible (image, text) pairings across a batch actually occurred
                \item \textbf{Extension:} Combines Resnet and ViT for the visual encoding with Transformer for the Textual encoder  
                \item \textbf{Application:} Image/object classification
                \item \textbf{Date (of first known publication):} 02/2021
                \item \textbf{Num. Params:} N/A
                \item \textbf{Corpus:} WIT (WebImageText) - 400 million text,image pairs
                \item \textbf{License:} Open, MIT license
                \item \textbf{Lab:} OpenAI
            \end{itemize}     

\subsection{CM3}

            \begin{itemize}
                \item \textbf{Reference:} \href{https://arxiv.org/abs/2201.07520}{\shortcite{aghajanyan2022cm3}}
                \item \textbf{Link:} N/A
                \item \textbf{Family:} HTML 
                \item \textbf{Pretraining Architecture:} Decoder
                \item \textbf{Pretraining Task:} Causality-masked LM
                \item \textbf{Extension:} This is somewhat similar to HTML in its use of structured training data. However, it is a different architecture and uses causal masking, which makes the model predict, at the end of the sequence, an entire missing span of text. It also includes image input via Vector Quantized Variational Autoencoding (VQ-VAE) tokens.
                \item \textbf{Application:} Multimodal language model with the ability to do structured prompting, zero-shot captioning, image generation, and entity linking (via target text prediction of hyperlinks)
                \item \textbf{Date (of first known publication):} 01/2022
                \item \textbf{Num. Params:}13B (largest)
                \item \textbf{Corpus:} CC-News, English Wikipedia
                \item \textbf{License:} N/A
                \item \textbf{Lab:} Facebook
            \end{itemize}

\subsection{CTRL}

            \begin{itemize}
                \item \textbf{Reference:} \shortcite{keskar2019ctrl}
                \item \textbf{Link:} \url{https://huggingface.co/docs/transformers/model_doc/ctrl}
                \item \textbf{Family:} 
                \item \textbf{Pretraining Architecture:} Decoder
                \item \textbf{Pretraining Task:}
                \item \textbf{Extension:} model can generate text conditioned on control codes that specify domain, style, topics, dates, entities, relationships between entities, plot points, and task-related behavior  
                \item \textbf{Application:} Controllable text generation
                \item \textbf{Date (of first known publication):} 09/2019
                \item \textbf{Num. Params:}1.63B
                \item \textbf{Corpus:} 140 GB of text including: Wikipedia (En, De, Es, Fr), Project Gutenberg, 45 subreddits, OpenWebText2, Amazon Reviews, Europarl and UN data from WMT, question-answer pairs from ELI5, and the MRQA shared task3, which includes the Stanford Question Answering Dataset, NewsQA, TriviaQA, SearchQA, HotpotQA , and Natural Questions 
                \item \textbf{License:} Open, BSD-3-Clause license
                \item \textbf{Lab:} Salesforce
            \end{itemize}
            
\subsection{DALL-E}

            \begin{itemize}
                \item \textbf{Reference:} \shortcite{ramesh2021zero}
                \item \textbf{Link:} \url{https://openai.com/blog/dall-e}
                \item \textbf{Family:} GPT 
                \item \textbf{Pretraining Architecture:} Decoder
                \item \textbf{Pretraining Task:} Caption prediction
                \item \textbf{Extension:} A differential variational auto-encoder is used to learn the visual codebook. The Transformer is a variation of GPT-3  
                \item \textbf{Application:} Text to image
                \item \textbf{Date (of first known publication):} 01/2021
                \item \textbf{Num. Params:}12B
                \item \textbf{Corpus:} 250 million text-images pairs from the internet
                \item \textbf{License:} N/A
                \item \textbf{Lab:} OpenAI
            \end{itemize}
            
\subsection{DALL-E 2}

            \begin{itemize}
                \item \textbf{Reference:} \shortcite{ramesh2022hierarchical}
                \item \textbf{Link:} \url{https://openai.com/dall-e-2}
                \item \textbf{Family:} CLIP, GLIDE 
                \item \textbf{Pretraining Architecture:} Encoder/Decoder
                \item \textbf{Pretraining Task:} Caption prediction
                \item \textbf{Extension:} Combines CLIP encoder and Diffusion decoder similar to GLIDE  
                \item \textbf{Application:} Text to image
                \item \textbf{Date (of first known publication):} 04/2022
                \item \textbf{Num. Params:}3.5B
                \item \textbf{Corpus:} Combination of the DALL-E and CLIP datasets
                \item \textbf{License:} Closed source, accessible through API
                \item \textbf{Lab:} OpenAI
            \end{itemize}

\subsection{DeBERTa}
            \begin{itemize}
                \item \textbf{Reference:} \shortcite{he2021deberta}
                \item \textbf{Link:} \url{https://huggingface.co/microsoft/deberta-large}
                \item \textbf{Family:} BERT 
                \item \textbf{Pretraining Architecture:} Encoder
                \item \textbf{Pretraining Task:} MLM
                \item \textbf{Extension:} Separate positional embedding vector independent from the content embedding using disentangled attention matrices for contents and relative positions
                \item \textbf{Application:} Same as BERT
                \item \textbf{Date (of first known publication):} 06/2020
                \item \textbf{Num. Params:} 134M (base), 384M (large), 750M (xlarge)
                \item \textbf{Corpus:} English Wikipedia, BookCorpus, OPENWEBTEXT and STORIES 
                \item \textbf{License:} Open, MIT license
                \item \textbf{Lab:} Microsoft
            \end{itemize}
   
\subsection{Decision Transformers}

            \begin{itemize}
                \item \textbf{Reference:} \href{https://arxiv.org/abs/2106.01345}{\shortcite{chen2021decision}}
                \item \textbf{Link:} \url{https://github.com/kzl/decision-transformer}
                \item \textbf{Family:} GPT, Control Transformers” (not per se a family, but grouping here those Transformers that try to model more general control, RL-like, tasks) 
                \item \textbf{Pretraining Architecture:} Decoder
                \item \textbf{Pretraining Task:} Next action prediction
                \item \textbf{Extension:} Decision Transformers use a GPT architecture and extend it by encoding trajectories in a way that they can be learned by an auto-regressive task  
                \item \textbf{Application:} General RL (reinforcement learning tasks)
                \item \textbf{Date (of first known publication):} 06/2021
                \item \textbf{Num. Params:}Same as GPT
                \item \textbf{Corpus:} Different corpus for different experiments
                \item \textbf{License:} Open, MIT license
                \item \textbf{Lab:} Google/UC Berkeley/Facebook
            \end{itemize}

\subsection{DialoGPT}
            \begin{itemize}
                \item \textbf{Reference:} \shortcite{zhang2019dialogpt}
                \item \textbf{Link:} \url{https://huggingface.co/docs/transformers/model_doc/dialogpt}
                \item \textbf{Family:} GPT 
                \item \textbf{Pretraining Architecture:} Decoder
                \item \textbf{Pretraining Task:} LM
                \item \textbf{Extension:} GPT-2 architecture trained on dialog data  
                \item \textbf{Application:} Text generation in dialog settings
                \item \textbf{Date (of first known publication):} 10/2019
                \item \textbf{Num. Params:}1.5B
                \item \textbf{Corpus:} 140M Reddit conversations
                \item \textbf{License:} Open, MIT license
                \item \textbf{Lab:} Microsoft
            \end{itemize}
            
\subsection{DistilBERT}
            \begin{itemize}
                \item \textbf{Reference:} \shortcite{sanh2019distilbert}
                \item \textbf{Link:} \url{https://huggingface.co/docs/transformers/model_doc/distilbert}
                \item \textbf{Family:} BERT 
                \item \textbf{Pretraining Architecture:} Encoder
                \item \textbf{Pretraining Task:} MLM/NSP
                \item \textbf{Extension:} Compressed version of BERT using distillation, which is much more efficient given the same number of parameters  
                \item \textbf{Application:} Same as BERT
                \item \textbf{Date (of first known publication):} 10/2019
                \item \textbf{Num. Params:}66M
                \item \textbf{Corpus:} Same as BERT
                \item \textbf{License:} Open, Apache-2.0
                \item \textbf{Lab:} Huggingface
            \end{itemize}
            
\subsection{DQ-BART}

            \begin{itemize}
                \item \textbf{Reference:} \shortcite{li2022dq}
                \item \textbf{Link:} \url{https://github.com/amazon-science/dq-bart}
                \item \textbf{Family:} BART 
                \item \textbf{Pretraining Architecture:} Encoder/Decoder
                \item \textbf{Pretraining Task:} DAE
                \item \textbf{Extension:} Adds quantization and distillation to a BART model to improve performance and model size  
                \item \textbf{Application:} Text generation and understanding
                \item \textbf{Date (of first known publication):} 03/2022
                \item \textbf{Num. Params:}Up to 30x reduction in parameters compared to standard BART
                \item \textbf{Corpus:} CNN/DM, XSUM, ELI5, WMT16 En-Ro (~1M tokens)
                \item \textbf{License:} Open, Apache-2.0
                \item \textbf{Lab:} Amazon
            \end{itemize}

\subsection{Dolly}
            \begin{itemize}
                \item \textbf{Reference:} See blog post\footnote{\url{https://www.databricks.com/blog/2023/04/12/dolly-first-open-commercially-viable-instruction-tuned-llm}}
                \item \textbf{Link:} \url{https://huggingface.co/databricks/dolly-v1-6b}
                \item \textbf{Family:} GPT
                \item \textbf{Pretraining Architecture:} Decoder
                \item \textbf{Fine-tuning Task:} human instructions
                \item \textbf{Extension:} fine-tuned based on the GPT-J-6B (V1) and Pythia model (V2)
                \item \textbf{Application:} Similar to Alpaca
                \item \textbf{Date (of first known publication):} 03/2023
                \item \textbf{Num. Params:} V1: 6B, V2: 12B
                \item \textbf{Corpus:} V1: Instruction corpus same as Alpaca, V2: databricks own dataset.
                \item \textbf{License:} Open
                \item \textbf{Lab:} Databricks, Inc
            \end{itemize}

\subsection{E5}
            \begin{itemize}
                \item \textbf{Reference:} \shortcite{wang2022e5}
                \item \textbf{Link:} \url{https://huggingface.co/intfloat/e5-large}
                \item \textbf{Family:} BERT 
                \item \textbf{Pretraining Architecture:} Encoder
                \item \textbf{Fine-tuning Task:} Semantic similarity using contrastive loss
                \item \textbf{Extension:} Fine-tunes BERT-based models to create text string embeddings optimized for semantic relatedness.  
                \item \textbf{Application:} Text embeddings for semantic relatedness tasks such as text clustering or search retrieval.
                \item \textbf{Date (of first known publication):} 12/2022
                \item \textbf{Num. Params:} 300M (large version)
                \item \textbf{Corpus:} MS-MARCO, NQ, NLI
                \item \textbf{License:} Open, MIT license
                \item \textbf{Lab:} Microsoft
            \end{itemize}

\subsection{ELECTRA}
            \begin{itemize}
                \item \textbf{Reference:}~\shortcite{clark2020electra}
                \item \textbf{Link:} \url{https://huggingface.co/docs/transformers/model_doc/electra}
                \item \textbf{Family:} BERT
                \item \textbf{Pretraining Architecture:} Encoder
                \item \textbf{Pretraining Task:} RTD
                \item \textbf{Extension:} Applied new training techniques including Replaced Token Detection  
                \item \textbf{Application:} 03/2020
                \item \textbf{Date (of first known publication):} 2020
                \item \textbf{Num. Params:}Base = 110M, Large = 330M
                \item \textbf{Corpus:} Same as BERT except for Large which is same as XLNet
                \item \textbf{License:} Open, Apache-2.0
                \item \textbf{Lab:} Stanford/Google
            \end{itemize}

\subsection{ERNIE}

            \begin{itemize}
                \item \textbf{Reference:}\href{https://arxiv.org/abs/1905.07129}{\shortcite{zhang2019ernie}}
                \item \textbf{Link:} N/A
                \item \textbf{Family:} BERT 
                \item \textbf{Pretraining Architecture:} Encoder
                \item \textbf{Pretraining Task:} MLM
                \item \textbf{Extension:} Uses BERT for Encoder architecture, but stacks and aggregates two of them for text and entities. This architecture could be understood as BERT for text + knowledge graphs  
                \item \textbf{Application:} Knowledge intensive related tasks that might benefit from knowledge graphs or entities such as entity recognition
                \item \textbf{Date (of first known publication):} 05/2019
                \item \textbf{Num. Params:} Ernie-ViLG 2.0 = 10B, Ernie 3.0 Titan = 260B
                \item \textbf{Corpus:} English Wikipedia + Wikidata for entitites (note that they initialize model to original BERT parameter values
                \item \textbf{License:} Closed source
                \item \textbf{Lab:} Baidu, Pengcheng Lab
            \end{itemize} 
            
\subsection{Flamingo}

            \begin{itemize}
                \item \textbf{Reference:} \href{https://arxiv.org/abs/2204.14198}{\shortcite{alayrac2022flamingo}}
                \item \textbf{Link:} N/A
                \item \textbf{Family:} Chinchilla 
                \item \textbf{Pretraining Architecture:} Decoder
                \item \textbf{Pretraining Task:} Log likelihood of text given some visual input
                \item \textbf{Extension:} It uses a frozen textual language model (like Chinchilla) conditioned on the visual representation, which is encoded from a Normalizer-Free ResNet  
                \item \textbf{Application:} Text to image
                \item \textbf{Date (of first known publication):} 04/2022
                \item \textbf{Num. Params:}80B (largest)
                \item \textbf{Corpus:} MultiModal MassiveWeb (M3W): 185 million images and 182 GB text + a number of text paired with image datasets: ALIGN + LTIP (Long Text \& Image Pairs) = 312 million images, and VTP (Video \& Text Pairs) = 27 million short videos (approximately 22 seconds on average)
                \item \textbf{License:} Closed source
                \item \textbf{Lab:} Deepmind
            \end{itemize}

\subsection{Flan-T5}

            \begin{itemize}
                \item \textbf{Reference:} \href{https://arxiv.org/abs/2210.11416}{\shortcite{hyung2022flant5}}
                \item \textbf{Link:} \url{https://huggingface.co/docs/transformers/model_doc/flan-t5}
                \item \textbf{Family:} T5 
                \item \textbf{Pretraining Architecture:} Encoder/Decoder
                \item \textbf{Fine-tuning Task:} Instructions for zero-shot and few-shot tasks
                \item \textbf{Extension:} Flan-T5 is generated by "Flan Finetuning" the T5 models: (1) scaling the number of tasks to 1,836, (2) scaling the model size, and (3) finetuning on chain-of-thought data.
                \item \textbf{Application:} The primary use is to understand how to improve large language models with the right kind of instruction fine-tuning. The focus is research on zero-shot and in-context few-shot learning NLP tasks, such as reasoning, and question answering; advancing fairness and safety research, and understanding limitations of current large language models
                \item \textbf{Date (of first known publication):} 11/2022
                \item \textbf{Num. Params:} 80M(small), 250M(base), 780M(large), 3B(xl), 11B(xxl)
                \item \textbf{Corpus:} Flan finetuned with tasks in Muffin, T0-SF, NIV2, and CoT.
                \item \textbf{License:} Open, Apache-2.0
                \item \textbf{Lab:} Google
            \end{itemize}

\subsection{Flan-PaLM}
            \begin{itemize}
                \item \textbf{Reference:} \href{https://arxiv.org/abs/2210.11416}{ \shortcite{hyung2022flant5}}
                \item \textbf{Link:} N/A
                \item \textbf{Family:} PaLM 
                \item \textbf{Pretraining Architecture:} Decoder
                \item \textbf{Fine-tuning Task:} Instructions for zero-shot and few-shot tasks
                \item \textbf{Extension:} Flan-PaLM is generated by "Flan Finetuning" the PaLM models: (1) scaling the number of tasks to 1,836, (2) scaling the model size, and (3) finetuning on chain-of-thought data.
                \item \textbf{Application:} Same as Flan-T5. The goal is to show Flan finetuning can even improve on the largest Google LMs (+9.4\% improvement average across tasks), with improvements to chain of thought, self consistency, multilingual tasks, arithmetic reasoning
                \item \textbf{Date (of first known publication):} 11/2022
                \item \textbf{Num. Params:} 8B, 62B, 540B
                \item \textbf{Corpus:} Flan finetuned with tasks in Muffin, T0-SF, NIV2, and CoT.
                \item \textbf{License:} Closed source
                \item \textbf{Lab:} Google
            \end{itemize}

\subsection{Galactica}

            \begin{itemize}
                \item \textbf{Reference:} \href{https://arxiv.org/abs/2211.09085}{\shortcite{taylor2022galactica}}
                \item \textbf{Link:} \url{https://galactica.org}
                \item \textbf{Family:} Transformer 
                \item \textbf{Pretraining Architecture:} Decoder
                \item \textbf{Pretraining Task:} LM for scientific domain
                \item \textbf{Extension:} Transformer based architecture in a decoder-only setup with a few modifications. Data extensions include special tokens for working memory, citations, genetic data, and a few other biology related tasks. 
                \item \textbf{Application:} The models are designed to perform scientific tasks, including but not limited to citation prediction, scientific QA, mathematical reasoning, summarization, document generation, molecular property prediction and entity extraction.
                \item \textbf{Date (of first known publication):} 11/2022
                \item \textbf{Num. Params:} mini: 125M, base: 1.3B, standard: 6.7B, large: 30B, huge: 120B
                \item \textbf{Corpus:} Trained on 106 billion tokens of open-access scientific text and data. This includes papers, textbooks, scientific websites, encyclopedias, reference material, knowledge bases, and more
                \item \textbf{License:} Limited, non-commerical CC BY-NC 4.0 license
                \item \textbf{Lab:} Meta
            \end{itemize}

\subsection{Gato}

            \begin{itemize}
                \item \textbf{Reference:} \shortcite{reed2022generalist}
                \item \textbf{Link:} \url{https://www.deepmind.com/blog/a-generalist-agent}
                \item \textbf{Family:} “Control Transformers” (not per se a family, but grouping here those Transformers that try to model more general control, RL-like, tasks) 
                \item \textbf{Pretraining Architecture:} Decoder
                \item \textbf{Pretraining Task:} MLM (where tokens are either text or agent actions)
                \item \textbf{Extension:} The standard decoder-only Transformer architecture is preceded by an embedding layer that can embed text and images, plus add position encodings to add spatial information when applicable.  
                \item \textbf{Application:} Gato presents a generalizable agent that can be used beyond text to tasks such as playing Atari or controlling a robot arm.
                \item \textbf{Date (of first known publication):} 05/2022
                \item \textbf{Num. Params:}1.2B
                \item \textbf{Corpus:} 1.5T tokens including standard text (e.g. MassiveText), vision (e.g. ALIGN), and simulation environments (e.g. ALE Atari, or RGB Stacking Real Robot)
                \item \textbf{License:} Closed source
                \item \textbf{Lab:} Deepmind
            \end{itemize}

\subsection{GLaM}

            \begin{itemize}
                \item \textbf{Reference:} \shortcite{du2022glam}
                \item \textbf{Link:} See blog post\footnote{\url{https://ai.googleblog.com/2021/12/more-efficient-in-context-learning-with.html}}
                \item \textbf{Family:} Transformer 
                \item \textbf{Pretraining Architecture:} Decoder
                \item \textbf{Pretraining Task:} LM
                \item \textbf{Extension:} GLaM introduces a Mixture of 64 Experts to increase parameter count and generalization properties in a somewhat standard decoder-only. Transformer architecture. Only two experts get activated at a time per token, which makes the model also more efficient in training and inference.  
                \item \textbf{Application:} General language modeling
                \item \textbf{Date (of first known publication):} 12/2021
                \item \textbf{Num. Params:}1.2T across 64 experts, but only 96B get activated for inference
                \item \textbf{Corpus:} 1.6T tokens including web pages filtered by Wikipedia and books for quality
                \item \textbf{License:} Closed source
                \item \textbf{Lab:} Google
            \end{itemize}

\subsection{GLIDE}

            \begin{itemize}
                \item \textbf{Reference:} \href{https://arxiv.org/abs/2112.10741}{\shortcite{nichol2021glide}}
                \item \textbf{Link:} \url{https://github.com/openai/glide-text2im}
                \item \textbf{Family:} Diffusion models 
                \item \textbf{Pretraining Architecture:} Encoder
                \item \textbf{Pretraining Task:}  Caption prediction
                \item \textbf{Extension:} GLIDE can be seen as an extension of the ADM (Ablated Diffusion Model) by the same authors. However, ADM is not per se a Transformer architecture although it does resemble one in some of the configurations the authors use. Given that ADM is by the same authors and was quickly followed up by GLIDE, we think it is fair to consider GLIDE as the first of its kind.  
                \item \textbf{Application:} Text to image
                \item \textbf{Date (of first known publication):} 12/2021
                \item \textbf{Num. Params:}3.5B diffusion model (2.3B for visual encoding, 1.2B for textual) + 1.5B for model for upsampling
                \item \textbf{Corpus:} Same as DALL-E
                \item \textbf{License:} Open, MIT license
                \item \textbf{Lab:} OpenAI
            \end{itemize}

\subsection{GLM}

            \begin{itemize}
                \item \textbf{Reference:} \href{https://arxiv.org/abs/2103.10360}{\shortcite{zhengxiao2022glm}}
                \item \textbf{Link:} \url{https://github.com/THUDM/GLM-130B}
                \item \textbf{Family:} GLM (General Language Model)
                \item \textbf{Pretraining Architecture:} Encoder and decoder
                \item \textbf{Pretraining Task:} Auto regressive blank infilling
                \item \textbf{Extension:} GLM has a bidirectional encoder and a unidirectional decoder in a unified model.
                \item \textbf{Application:} a General Language Model pretrained with an autoregressive blank-filling objective and can be finetuned on various natural language understanding and generation tasks.
                \item \textbf{Date (of first known publication):} 03/2022
                \item \textbf{Num. Params:} Base = 110M, Large = 335M, and also 2B, 10B, 130B
                \item \textbf{Corpus:} Pile, GLM-130B Chinese corpora, P3, DeepStruct finetuning dataset
                \item \textbf{License:} Open, MIT license
                \item \textbf{Lab:} Tsinghua
            \end{itemize}
            
\subsection{Global Context ViT}

            \begin{itemize}
                \item \textbf{Reference:} \href{https://arxiv.org/abs/2206.09959}{\shortcite{hatamizadeh2022global}}
                \item \textbf{Link:} \url{https://github.com/NVlabs/GCVit}
                \item \textbf{Family:} ViT 
                \item \textbf{Pretraining Architecture:} Encoder
                \item \textbf{Pretraining Task:} Image classification
                \item \textbf{Extension:} hierarchical ViT architecture consisting of local and global self-attention modules  
                \item \textbf{Application:} Image generation
                \item \textbf{Date (of first known publication):} 06/2022
                \item \textbf{Num. Params:} 90M
                \item \textbf{Corpus:} Imagenet-1K and other task dependent dataasets
                \item \textbf{License:} Limited, non-commercial license CC-BY-NC-SA-4.0
                \item \textbf{Lab:} NVidia
            \end{itemize}

\subsection{Gopher}

            \begin{itemize}
                \item \textbf{Reference:} \shortcite{rae2021scaling}
                \item \textbf{Link:} See blog post\footnote{\url{https://www.deepmind.com/blog/language-modelling-at-scale-gopher-ethical-considerations-and-retrieval}}
                \item \textbf{Family:} GPT 
                \item \textbf{Pretraining Architecture:} Decoder
                \item \textbf{Pretraining Task:} LM
                \item \textbf{Extension:} Same as GPT-2 but use RSNorm instead of LayerNorm and relative positional encoding rather than absolute  
                \item \textbf{Application:} Mostly Language Modeling and NLU, but also extensible like GPT
                \item \textbf{Date (of first known publication):} 12/2021
                \item \textbf{Num. Params:}280B
                \item \textbf{Corpus:} Massive Text (2.35 billion documents, or about 10.5 TB of text including Massive Web, Books, Github, News, C4, and Wikipedia.
                \item \textbf{License:} Closed source
                \item \textbf{Lab:} Deepmind
            \end{itemize}

\subsection{GopherCite}

            \begin{itemize}
                \item \textbf{Reference:} \href{https://arxiv.org/abs/2203.11147}{\shortcite{menick2022teaching}}
                \item \textbf{Link:} See blog post\footnote{\url{https://www.deepmind.com/blog/gophercite-teaching-language-models-to-support-answers-with-verified-quotes}}
                \item \textbf{Family:} GPT 
                \item \textbf{Pretraining Architecture:} Decoder
                \item \textbf{Pretraining Task:} LM
                \item \textbf{Extension:} GopherCite is based on Gopher but adds a step using RLHP (Reinforcement Learning from Human Preferences) to learn whether not only a response is plausible but also supported  
                \item \textbf{Application:} Dialog systems, Q\&A, general language generation tasks
                \item \textbf{Date (of first known publication):} 03/2022
                \item \textbf{Num. Params:}280B
                \item \textbf{Corpus:} Same as Gopher plus specific dataset generated in the RLHP process
                \item \textbf{License:} Closed source
                \item \textbf{Lab:} Deepmind
            \end{itemize}

\subsection{GPT}

            \begin{itemize}
                \item \textbf{Reference:} \shortcite{radford2018improving}
                \item \textbf{Link:} \url{https://huggingface.co/docs/transformers/model_doc/openai-gpt}
                \item \textbf{Family:} GPT 
                \item \textbf{Pretraining Architecture:} Decoder
                \item \textbf{Pretraining Task:} LM
                \item \textbf{Extension:}   
                \item \textbf{Application:} Text generation, but adaptable to many other NLP tasks when fine tuned.
                \item \textbf{Date (of first known publication):} 06/2018
                \item \textbf{Num. Params:}117M
                \item \textbf{Corpus:} Unsupervised Pretraining on BookCorpus dataset. Supervised Finetuning on several task-specific datasets including SNLI, RACE, Quora…
                \item \textbf{License:} N/A
                \item \textbf{Lab:} OpenAI
            \end{itemize}

\subsection{GPT-2}

            \begin{itemize}
                \item \textbf{Reference:} \shortcite{radford2019language}
                \item \textbf{Link:} \url{https://huggingface.co/docs/transformers/model_doc/gpt2}
                \item \textbf{Family:} GPT 
                \item \textbf{Pretraining Architecture:} Decoder
                \item \textbf{Pretraining Task:} LM
                \item \textbf{Extension:} Minor extensions to the GPT architecture (e.g. layer normalization moved to the input of each sub-layer, or increased context size from 512 to 1024)  
                \item \textbf{Application:} Text generation, but adaptable to many other NLP tasks when fine tuned.
                \item \textbf{Date (of first known publication):} 02/2019
                \item \textbf{Num. Params:} 124M, 355M, 774M, 1.5B
                \item \textbf{Corpus:} 8 million web pages (40 GB). 10X GPT . WebText dataset is created by crawling all links at Reddit with at least 3 Karma points.
                \item \textbf{License:} Open, Modified MIT license
                \item \textbf{Lab:} OpenAI
            \end{itemize}   

\subsection{GPT-3}

            \begin{itemize}
                 \item \textbf{Reference:} \shortcite{brown2020language}
                 \item \textbf{Link:} \url{https://github.com/openai/gpt-3}
                 \item \textbf{Family:} GPT 
                \item \textbf{Pretraining Architecture:} Decoder
                \item \textbf{Pretraining Task:} LM
                \item \textbf{Extension:} Same as GPT-2 with the only addition of alternating dense and locally banded sparse attention patterns, inspired by the Sparse Transformer
                \item \textbf{Application:} Initially text generation, but has over time been used for a large range of applications in areas such as code generation, but also image and audio generation
                \item \textbf{Date (of first known publication):} 05/2020
                \item \textbf{Num. Params:}175 B
                \item \textbf{Corpus:} 500B tokens including CommonCrawl (410B), WebText2 (19B), Books1 (12B), Books2 (55B), and Wikipedia (3B)
                \item \textbf{License:} Closed source
                \item \textbf{Lab:} OpenAI
            \end{itemize}

\subsection{GPT-3.5}

            \begin{itemize}
                \item \textbf{Reference:} N/A
                \item \textbf{Link:} \url{https://platform.openai.com/docs/model-index-for-researchers/models-referred-to-as-gpt-3-5}
                \item \textbf{Family:} GPT 
                \item \textbf{Pretraining Architecture:} Decoder
                \item \textbf{Pretraining Task:} LM
                \item \textbf{Extension:} The GPT3.5 series includes a number of models like Davinci-003. They are basically versions of the InstructGPT model. See blog post\footnote{\url{https://scale.com/blog/gpt-3-davinci-003-comparison}} for details on the comparison of the performance to older GPT3 models.  
                \item \textbf{Application:} Dialog and general language, but there is a code specific model - codex
                \item \textbf{Date (of first known publication):} 10/2022
                \item \textbf{Num. Params:}175B
                \item \textbf{Corpus:} Same as InstructGPT
                \item \textbf{License:} Closed source, accessible through API
                \item \textbf{Lab:} OpenAI
            \end{itemize}
            
\subsection{GPT-J}
            \begin{itemize}
                \item \textbf{Reference:} \shortcite{gpt-j}
                \item \textbf{Link:} \url{https://huggingface.co/EleutherAI/gpt-j-6B}
                \item \textbf{Family:} GPT 
                \item \textbf{Pretraining Architecture:} Decoder
                \item \textbf{Pretraining Task:} LM
                \item \textbf{Extension:} GPT-J 6B is a Transformer model trained using Mesh Transformer JAX and same tokenizer as GPT2/3  
                \item \textbf{Application:} Same as GPT-3
                \item \textbf{Date (of first known publication):} 05/2021
                \item \textbf{Num. Params:} 6B
                \item \textbf{Corpus:} Pile corpus, a large-scale curated dataset created by EleutherAI.
                \item \textbf{License:} Open, Apache-2.0
                \item \textbf{Lab:} EleutherAI
            \end{itemize}
            
\subsection{GPT-Neo}

            \begin{itemize}
                \item \textbf{Reference:} \url{https://huggingface.co/docs/transformers/model_doc/gpt_neo}
                \item \textbf{Link:} \url{https://github.com/EleutherAI/gpt-neo}
                \item \textbf{Family:} GPT 
                \item \textbf{Pretraining Architecture:} Decoder
                \item \textbf{Pretraining Task:} LM
                \item \textbf{Extension:} Similar to GPT-2 but uses local attention in every other layer with a window size of 256 tokens  
                \item \textbf{Application:} Text generation, but adaptable to many other NLP tasks when fine tuned
                \item \textbf{Date (of first known publication):} 03/2021
                \item \textbf{Num. Params:} 5B, 2.7B (XL)
                \item \textbf{Corpus:} Pile — 840 GB open source text dataset that combines 22 pre existing datasets
                \item \textbf{License:} Open, MIT license
                \item \textbf{Lab:} EleutherAI
            \end{itemize}
 
\subsection{GPT-NeoX-20B}

            \begin{itemize}
                \item \textbf{Reference:} \href{https://arxiv.org/abs/2204.06745}{\shortcite{black2022gpt}}
                \item \textbf{Link:} \url{https://huggingface.co/EleutherAI/gpt-neox-20b}
                \item \textbf{Family:} GPT 
                \item \textbf{Pretraining Architecture:} Decoder
                \item \textbf{Pretraining Task:} LM
                \item \textbf{Extension:} Similar to GPT-3 with rotary encoders instead of positional, parallel attention and feed forward layers, different initialization, and all dense layers instead of alternate dense/sparse  
                \item \textbf{Application:} same as GPT-3
                \item \textbf{Date (of first known publication):} 04/2022
                \item \textbf{Num. Params:}20B
                \item \textbf{Corpus:} Pile — 840 GB open source text dataset that combines 22 pre existing datasets
                \item \textbf{License:} Open, Apache-2.0
                \item \textbf{Lab:} EleutherAI
            \end{itemize}

\subsection{HTLM}

            \begin{itemize}
                \item \textbf{Reference:} \href{https://arxiv.org/abs/2107.06955}{\shortcite{aghajanyan2021htlm}}
                \item \textbf{Link:} N/A
                \item \textbf{Family:} BART 
                \item \textbf{Pretraining Architecture:} Encoder/Decoder
                \item \textbf{Pretraining Task:} DAE
                \item \textbf{Extension:} As opposed to BART, they don’t do sentence shuffling  
                \item \textbf{Application:} General purpose language model that allows structured HTML prompting 
                \item \textbf{Date (of first known publication):} 07/2021
                \item \textbf{Num. Params:}400M
                \item \textbf{Corpus:} 23TB of simplified HTML extracted from CommonCrawl
                \item \textbf{License:} N/A
                \item \textbf{Lab:} Facebook
            \end{itemize}

\subsection{Imagen}

            \begin{itemize}
                \item \textbf{Reference:} \shortcite{saharia2022photorealistic}
                \item \textbf{Link:} \url{https://imagen.research.google}
                \item \textbf{Family:} T5, CLIP, Diffusion models 
                \item \textbf{Pretraining Architecture:} T5 (or CLIP or BERT) for frozen text encoder + U-net architecture for cascaded diffusion models for text to image
                \item \textbf{Pretraining Task:} image/text pair prediction
                \item \textbf{Extension:} Imagen adds a few extensions to the U-net diffusion architecture (pooled embedding vector, cross attention over text embeddings, and Layer Normalizations)  
                \item \textbf{Application:} Text to image
                \item \textbf{Date (of first known publication):} 06/2022
                \item \textbf{Num. Params:}2B
                \item \textbf{Corpus:} a combination of internal datasets, with 460M image-text pairs, and the publicly available Laion dataset, with 400M image-text pairs
                \item \textbf{License:} Closed source
                \item \textbf{Lab:} Google
            \end{itemize}

\subsection{InstructGPT}
            \begin{itemize}
                \item \textbf{Reference:} \shortcite{ouyang2022training}
                \item \textbf{Link:} \url{https://github.com/openai/following-instructions-human-feedback}
                \item \textbf{Family:} GPT 
                \item \textbf{Pretraining Architecture:} Decoder
                \item \textbf{Pretraining Task:} LM
                \item \textbf{Extension:} GPTInstruct starts off with a pretrained GPT3 model and adds reward modeling through reinforcement learning after a supervised finetuning  
                \item \textbf{Application:} Knowledge-intensive dialog or language tasks
                \item \textbf{Date (of first known publication):} 01/2022
                \item \textbf{Num. Params:} Same as GPT3
                \item \textbf{Corpus:} Same as GPT3 for pretraining, but finetuned and optimized using labeler data and prompts
                \item \textbf{License:} Closed source, Accessible through API
                \item \textbf{Lab:} OpenAI
            \end{itemize}
            
\subsection{InstructOR}

            \begin{itemize}
                \item \textbf{Reference:} \shortcite{su2022instructor}
                \item \textbf{Link:} \url{https://huggingface.co/hkunlp/instructor-xl}
                \item \textbf{Family:} T5 
                \item \textbf{Pretraining Architecture:} Encoder/Decoder
                 \item \textbf{Fine-tuning Tasks:} Wide variety of instruction based text-to-text tasks 
                \item \textbf{Extension:} Fine-tunes T5 explicitly to optimize encoder to produce a general purpose text string embedding useful for many NLU tasks.
                \item \textbf{Application:} Any NLU task requiring a single text string embedding. As of April 2023 InstructOR is the top-ranked system on the Massive Text Embedding Benchmark (MTEB).\footnote{https://huggingface.co/spaces/mteb/leaderboard}
                \item \textbf{Date (of first known publication):} 12/2022
                \item \textbf{Num. Params:} 330M
                \item \textbf{Corpus:} Finetuned on MEDI 
                \item \textbf{License:} Open, Apache-2.0
                \item \textbf{Lab:} University of Hong Kong, University of Washington, META AI
            \end{itemize}

\subsection{Jurassic-1}

            \begin{itemize}
                \item \textbf{Reference:} \href{https://uploads-ssl.webflow.com/60fd4503684b466578c0d307/61138924626a6981ee09caf6_jurassic_tech_paper.pdf}{\shortcite{lieber2021jurassic}}
                \item \textbf{Link:} \url{https://github.com/ai21labs/lm-evaluation}
                \item \textbf{Family:} GPT 
                \item \textbf{Pretraining Architecture:} Decoder
                \item \textbf{Pretraining Task:} LM
                \item \textbf{Extension:} Very similar to GPT-3, but far more parameters and improved training efficiency mostly because of the improved tokenizer. Also, different ratio of depth to breadth  
                \item \textbf{Application:} Similar to GPT-3
                \item \textbf{Date (of first known publication):} 09/2021
                \item \textbf{Num. Params:} 178B (Jumbo), 17B (Grande), 7.5B (Large)
                \item \textbf{Corpus:} 300B tokens (same as GPT-3)
                \item \textbf{License:} Closed source, accessible through API
                \item \textbf{Lab:} AI21
            \end{itemize}
            
\subsection{LAMDA}

            \begin{itemize}
                \item \textbf{Reference:} \shortcite{thoppilan2022lamda}
                \item \textbf{Link:} See blog post\footnote{\url{https://ai.googleblog.com/2022/01/lamda-towards-safe-grounded-and-high.html}}
                \item \textbf{Family:} Transformer 
                \item \textbf{Pretraining Architecture:} Decoder
                \item \textbf{Pretraining Task:} LM
                \item \textbf{Extension:} LAMDA focuses on how to improve safety, quality, and groundeness using different fine-tuning strategies  
                \item \textbf{Application:} General language modeling, such as translation, summarization, question and answers.
                \item \textbf{Date (of first known publication):} 01/2022
                \item \textbf{Num. Params:}137B
                \item \textbf{Corpus:} 1.56T words from public dialog data and other public web documents
                \item \textbf{License:} Closed source
                \item \textbf{Lab:} Google
            \end{itemize}
            
\subsection{LLaMA}

            \begin{itemize}
                \item \textbf{Reference:} \shortcite{touvron2023llama}
                \item \textbf{Link:} \url{https://huggingface.co/docs/transformers/main/model_doc/llama}
                \item \textbf{Family:} Transformer
                \item \textbf{Pretraining Architecture:} Decoder
                \item \textbf{Pretraining Task:} LM
                \item \textbf{Extension:} LLaMA uses a Transformer architecture, and with extensions: Pre-normalization, SwiGLU activations, RoPE embeddings,  reduced memory usage and runtime through efficient implementation of the causal multi-head attention, checkpointing to reduce the amount of activations that are recomputed during the backward pass, model and sequence parallelism to reduce memory usage of the model, and uses 1.4T BPE tokens after tokenization.
                \item \textbf{Application:} Zero and few shot Commonsense reasoning, Question answering, Code generation and Reading comprehension.
                \item \textbf{Date (of first known publication):} 02/2023
                \item \textbf{Num. Params:} 7B, 13B, 33B and 65B
                \item \textbf{Corpus:} English CommonCrawl + C4 + Github + Wikipedia + Gutenberg and Books3 + ArXiv + Stack Exchange 
                \item \textbf{License:} Limited, Non-commercial bespoke license
                \item \textbf{Lab:} Meta
            \end{itemize}
            
\subsection{mBART}

            \begin{itemize}
                \item \textbf{Reference:} \shortcite{liu2020multilingual}
                \item \textbf{Link:} \url{https://huggingface.co/docs/transformers/model_doc/mbart}
                \item \textbf{Family:} BART 
                \item \textbf{Pretraining Architecture:} Encoder/Decoder
                \item \textbf{Pretraining Task:} DAE
                \item \textbf{Extension:} Extends BART to multilingual capability
                \item \textbf{Application:} Translation
                \item \textbf{Date (of first known publication):} 01/2020
                \item \textbf{Num. Params:} Same as BART
                \item \textbf{Corpus:} CC25 Corpus includes 25 monolingual corpuses in different languages. Largest corpuses are English (300 GB) and Russian (280GB)
                \item \textbf{License:} Open, MIT license
                \item \textbf{Lab:} facebook
            \end{itemize}

\subsection{Megatron}

            \begin{itemize}
                \item \textbf{Reference:} \shortcite{shoeybi2019megatron}
                \item \textbf{Link:} \url{https://github.com/NVIDIA/Megatron-LM}
                \item \textbf{Family:} GPT/BERT/T5 
                \item \textbf{Pretraining Architecture:} Encoder or Decorder, depending on the base model
                \item \textbf{Pretraining Task:} Same as base model
                \item \textbf{Extension:} Megatron is a family of models that extend previously known architectures (namely GPT-2 and BERT originally, but also T5 more recently) by introducing model parallelism primitives. In the case of BERT, the authors also replace the next sentence prediction head with sentence order prediction and use whole word n-gram masking.  
                \item \textbf{Application:} Same as base model
                \item \textbf{Date (of first known publication):} 03/2020
                \item \textbf{Num. Params:} 8.3B (GPT-like), 3.9B (BERT-like)
                \item \textbf{Corpus:} Original paper uses an aggregate dataset consisting of Wikipedia), CC-Stories), RealNews, and OpenWebtext
                \item \textbf{License:} Limited, Non-commercial usage
                \item \textbf{Lab:} NVidia
            \end{itemize}

\subsection{Minerva}

            \begin{itemize}
                \item \textbf{Reference:} \shortcite{lewkowycz2022solving}
                \item \textbf{Link:} See blog post\footnote{\url{https://ai.googleblog.com/2022/06/minerva-solving-quantitative-reasoning.html}}
                \item \textbf{Family:} PaLM 
                \item \textbf{Pretraining Architecture:} Decoder
                \item \textbf{Pretraining Task:} LM
                \item \textbf{Extension:} Extends PaLM by fine-tuning on the mathematical dataset  
                \item \textbf{Application:} Mathematical reasoning
                \item \textbf{Date (of first known publication):} 06/2022
                \item \textbf{Num. Params:}540B
                \item \textbf{Corpus:} Same as PaLM + 118GB dataset of scientific papers from the arXiv preprint server and web pages that contain mathematical expressions using LaTeX, MathJax, or other mathematical typesetting formats
                \item \textbf{License:} Closed source
                \item \textbf{Lab:} Google
            \end{itemize}

\subsection{MT-NLG (Megatron TuringNLG)}

            \begin{itemize}
                \item \textbf{Reference:} \shortcite{smith2022using}
                \item \textbf{Link:} See blog post\footnote{\url{https://developer.nvidia.com/blog/using-deepspeed-and-megatron-to-train-megatron-turing-nlg-530b-the-worlds-largest-and-most-powerful-generative-language-model/}}
                \item \textbf{Family:} GPT 
                \item \textbf{Pretraining Architecture:} Decoder
                \item \textbf{Pretraining Task:} LM
                \item \textbf{Extension:} Uses parallelization similar to Megatron to train a LM double the size of GPT-3  
                \item \textbf{Application:} Language generation and others (similar to GPT-3)
                \item \textbf{Date (of first known publication):} 10/2021
                \item \textbf{Num. Params:}530B
                \item \textbf{Corpus:} The Pile\footnote{\url{https://arxiv.org/abs/2101.00027}} (800GB dataset) + 2 Common Crawl snapshots
                \item \textbf{License:} Limited, Non-commercial usage
                \item \textbf{Lab:} NVidia
            \end{itemize}

\subsection{OpenAssistant LLaMa}
            \begin{itemize}
                \item \textbf{Reference:} N/A
                \item \textbf{Link:} \url{https://open-assistant.io/}
                \item \textbf{Family:} LLaMA
                \item \textbf{Pretraining Architecture:} Decoder
                \item \textbf{Extension:} Supervised fine-tuning on crowd sourced conversation/assistant data.
                \item \textbf{Application:} Same as ChatGPT, but open source. Compared to alternatives, it uses human generated conversation data
                \item \textbf{Date (of first known publication):} 04/2023
                \item \textbf{Num. Params:} 30B for LLaMa
                \item \textbf{Corpus:} Conversations collected by volunteers \shortcite{kopf2022openassistant} available at \url{https://huggingface.co/datasets/OpenAssistant/oasst1}
                \item \textbf{License:} Limited, Non-commercial bespoke license. There is also a version based on Pythia which is Apache licensed.
                \item \textbf{Lab:} Various open source contributors
            \end{itemize}
            
\subsection{OPT}

            \begin{itemize}
                \item \textbf{Reference:} \shortcite{zhang2022opt}
                \item \textbf{Link:} See blog post\footnote{\url{https://ai.facebook.com/blog/democratizing-access-to-large-scale-language-models-with-opt-175b/}}
                \item \textbf{Family:} GPT 
                \item \textbf{Pretraining Architecture:} Decoder
                \item \textbf{Pretraining Task:} LM
                \item \textbf{Extension:} Basically same architecture as GPT-3 but with some training improvements introduced in Megatron-LM  
                \item \textbf{Application:} Same as GPT-3
                \item \textbf{Date (of first known publication):} 05/2022
                \item \textbf{Num. Params:} 175B (and other smaller versions)
                \item \textbf{Corpus:} 180B tokens = RoBERTa + the Pile + PushShift.io Reddit
                \item \textbf{License:} Limited, non-commercial license
                \item \textbf{Lab:} Facebook
            \end{itemize}
            
\subsection{PalM}

            \begin{itemize}
                \item \textbf{Reference:} \shortcite{chowdhery2022palm}
                \item \textbf{Link:} See blog post\footnote{\url{https://ai.googleblog.com/2022/04/pathways-language-model-palm-scaling-to.html}}
                \item \textbf{Family:} Transformer 
                \item \textbf{Pretraining Architecture:} Decoder
                \item \textbf{Pretraining Task:} LM
                \item \textbf{Extension:} Palm uses a typical decoder-only Transformer architecture, but adds quite a few extensions: SwiGLU activations, parallel layers, multi-query attention, RoPE embeddings, Shared Input-Output Embeddings, no biases, and a 256k SentencePiece vocabulary generated from the training data.  
                \item \textbf{Application:} PalM is designed as a general purpose language model with applicability to hundreds of different language tasks
                \item \textbf{Date (of first known publication):} 04/2022
                \item \textbf{Num. Params:} 540B
                \item \textbf{Corpus:} 780B tokens from filtered webpages, books, Wikipedia, news articles, source code, and social media conversations. Code includes 24 programming languages.
                \item \textbf{License:} Closed source, Accessible through API
                \item \textbf{Lab:} Google
            \end{itemize}

\subsection{Pegasus}

            \begin{itemize}
                \item \textbf{Reference:} \shortcite{zhang2020pegasus}
                \item \textbf{Link:} \url{https://huggingface.co/docs/transformers/model_doc/pegasus}
                \item \textbf{Family:}  Transformer
                \item \textbf{Pretraining Architecture:} Encoder/Decoder
                \item \textbf{Pretraining Task:} DAE (more concretely GSG) and MLM
                \item \textbf{Extension:} Extends vanilla Transformer by using a different pretraining task (GSG: Gap Sentence Generation) that is better suited for summarization  
                \item \textbf{Application:} Summarization
                \item \textbf{Date (of first known publication):} 12/2019
                \item \textbf{Num. Params:} Base = 223M, Large = 568M
                \item \textbf{Corpus:} C4 (750GB) + HugeNews (3.8 TB)
                \item \textbf{License:} N/A
                \item \textbf{Lab:} UCL/Google
            \end{itemize}

\subsection{Pythia}
            \begin{itemize}
                \item \textbf{Reference:} \href{https://arxiv.org/abs/2304.01373}{\shortcite{biderman2023pythia}}
                \item \textbf{Link:} \url{https://github.com/EleutherAI/pythia}
                \item \textbf{Family:} Pythia
                \item \textbf{Pretraining Architecture:} Decoder
                \item \textbf{Extension:} Trained with the library GPT-NeoX
                \item \textbf{Application:} Research on language model's behavior, functionality, and limitations.
                \item \textbf{Date (of first known publication):} 04/2023
                \item \textbf{Num. Params:} 70M, 160M, 410M, 1B, 1.4B, 2.8B, 6.9B, 12B
                \item \textbf{Corpus:} Pile
                \item \textbf{License:} Open, Apache-2.0
                \item \textbf{Lab:} Eleuther AI
            \end{itemize}
            
\subsection{RoBERTa}

            \begin{itemize}
                \item \textbf{Reference:} \shortcite{liu2019roberta}
                \item \textbf{Link:} \url{https://huggingface.co/docs/transformers/model_doc/roberta}
                \item \textbf{Family:} BERT 
                \item \textbf{Pretraining Architecture:} Encoder
                \item \textbf{Pretraining Task:} MLM (Dynamic)
                \item \textbf{Extension:} Extension of BERT with optimized training procedure and more data  
                \item \textbf{Application:} Same as BERT
                \item \textbf{Date (of first known publication):} 07/2019
                \item \textbf{Num. Params:} 356M
                \item \textbf{Corpus:} Same as BERT + CC News + OpenWebText + Stories (~33B Tokens)
                \item \textbf{License:} N/A
                \item \textbf{Lab:} UW/Google
            \end{itemize}
            
\subsection{SeeKer}

            \begin{itemize}
                \item \textbf{Reference:} \shortcite{shuster2022language}
                \item \textbf{Link:} \url{https://parl.ai/projects/seeker}
                \item \textbf{Family:} GPT (but can extend any family) 
                \item \textbf{Pretraining Architecture:} Encoder/decoder or decoder only, depending on the base model it’s extending
                \item \textbf{Pretraining Task:} LM training, Dialogue training
                \item \textbf{Extension:} SeeKer is an extension that can be applied to any Transformer architecture by introducing “search”, “knowledge”, and “response” modules that are introduced during pretraining  
                \item \textbf{Application:} Same as base models
                \item \textbf{Date (of first known publication):} 03/2022
                \item \textbf{Num. Params:} SeeKeR Dialogue: 400M, 3B; SeeKeR LM: 365M, 762M, 1.5B, R2C2 BlenderBot: 400M, 3B
                \item \textbf{Corpus:} Wizard of the Internet/Wikipedia, PersonaChat, Blended Skill Talk, Empatheic Dialogues, Multi-Session Chat, MS MARCO, Natural questions, SQuAD, TriviaQA
                \item \textbf{License:} the code is open sourced.
                \item \textbf{Lab:} Facebook
            \end{itemize}

\subsection{Sparrow}

            \begin{itemize}
                \item \textbf{Reference:} \href{https://arxiv.org/abs/2209.14375}{\shortcite{glaese2022improving}}
                \item \textbf{Link:} N/A
                \item \textbf{Family:} GPT 
                \item \textbf{Pretraining Architecture:} Decoder
                \item \textbf{Pretraining Task:} LM
                \item \textbf{Extension:} Starts from the Chinchilla 70B model but adds RLHF (Reinforcement Learning with Human Feedback). It also adds inline evidence a la GopherCite  
                \item \textbf{Application:} Dialog agents and general language generation applications like Q\&A
                \item \textbf{Date (of first known publication):} 09/2022
                \item \textbf{Num. Params:} 70B
                \item \textbf{Corpus:} Same as Chinchilla + interactive data gathering with human annotators during the RLHF process
                \item \textbf{License:} Closed source
                \item \textbf{Lab:} Deepmind
            \end{itemize}
 
\subsection{StableDiffusion}

            \begin{itemize}
                \item \textbf{Reference:} \shortcite{rombach2022high}
                \item \textbf{Link:} \url{https://huggingface.co/CompVis/stable-diffusion}
                \item \textbf{Family:} Diffusion 
                \item \textbf{Pretraining Architecture:} Encoder/Decoder
                \item \textbf{Pretraining Task:} Caption prediction
                \item \textbf{Extension:} Stable diffusion is basically the Latent Diffusion model developed by LMU Munich researchers + some learnings on conditional diffusion from DALL-e and Imagen  
                \item \textbf{Application:} Text to image 
                \item \textbf{Date (of first known publication):} 12/2021
                \item \textbf{Num. Params:} 890M (although there are different, smaller, variants)
                \item \textbf{Corpus:} LAION-5B, a publicly available dataset derived from Common Crawl
                \item \textbf{License:} open, CreativeML Open RAIL++-M License
                \item \textbf{Lab:} LMU Munich + Stability.ai + Eleuther.ai
            \end{itemize}

\subsection{Swin Transformer}

            \begin{itemize}
                \item \textbf{Reference:} \shortcite{liu2021swin}
                \item \textbf{Link:} \url{https://github.com/microsoft/Swin-Transformer}
                \item \textbf{Family:} ViT 
                \item \textbf{Pretraining Architecture:} Encoder
                \item \textbf{Pretraining Task:} Same as ViT
                \item \textbf{Extension:} Extends ViT by replacing the standard multi-head self attention (MSA) module by a module based on shifted windows (Swin) allowing ViT-like architectures to generalize to higher resolution images  
                \item \textbf{Application:} Image (object detection, image classification..)
                \item \textbf{Date (of first known publication):} 03/2021
                \item \textbf{Num. Params:} 29M-197M
                \item \textbf{Corpus:} Imagenet and Imagenet-22k
                \item \textbf{License:} the code is open sourced, with MIT-license
                \item \textbf{Lab:} Microsoft
            \end{itemize}

\subsection{Switch}

            \begin{itemize}
                \item \textbf{Reference:} \href{https://arxiv.org/abs/2101.03961}{\shortcite{fedus2021switch}}
                \item \textbf{Link:} \url{https://github.com/google-research/t5x}
                \item \textbf{Family:} T5 
                \item \textbf{Pretraining Architecture:} Encoder/Decoder
                \item \textbf{Pretraining Task:} DAE
                \item \textbf{Extension:} Goal to increase parameter count while keeping FLOP operations constant by using efficient routing of MoE (Mixture of Experts)  
                \item \textbf{Application:} General language tasks (e.g. question answering)
                \item \textbf{Date (of first known publication):} 01/2021
                \item \textbf{Num. Params:} 1T
                \item \textbf{Corpus:} Colossal Clean Crawled Corpus
                \item \textbf{License:} Open, Apache-2.0
                \item \textbf{Lab:} Google
            \end{itemize}

\subsection{T0}

            \begin{itemize}
                \item \textbf{Reference:} \href{https://arxiv.org/abs/2110.08207}{\shortcite{victor2022t0}}
                \item \textbf{Link:} \url{https://huggingface.co/bigscience/T0}
                \item \textbf{Family:} T5
                \item \textbf{Pretraining Architecture:} Encoder/Decoder
                \item \textbf{Fine-tuning Task:} Natural language prompts
                \item \textbf{Extension:} T0 stands for "T5 for Zero Shot", obtained by fine-tuning the T5 model on multitask mixture covering many different NLP tasks. Compared with T0, T0p and T0pp were fine-tuned with more datasets. T0pp is recommended as it leads (on average) to the best performances on a variety of NLP tasks.
                \item \textbf{Application:} Perform zero-shot inference tasks by specifying the query in natural language, and the models will generate a prediction.
                \item \textbf{Date (of first known publication):} 03/2022
                \item \textbf{Num. Params:} T0-3B: 3 billion, T0, T0p, T0pp: 11 billion
                \item \textbf{Corpus:} T0 (Multiple-choice QA, Extractive QA, Closed-Book QA, Structure-To-Text, Sentiment, Summarization, Topic Classification, Paraphrase Identification. T0p (same as T0, with additional datasets from GPT-3's evaluation suite). T0pp (same as T0p, with additional datasets from SuperGLUE, excluding NLI sets)
                \item \textbf{License:} Open, Apache-2.0
                \item \textbf{Lab:} BigScience
            \end{itemize}
            
\subsection{T5}

            \begin{itemize}
                \item \textbf{Reference:} \shortcite{raffel2020exploring}
                \item \textbf{Link:} \url{https://huggingface.co/docs/transformers/model_doc/t5}
                \item \textbf{Family:} Transformer
                \item \textbf{Pretraining Architecture:} Encoder/Decoder
                \item \textbf{Pretraining Task:} DAE
                \item \textbf{Extension:} Same as original Transformer with some additions such as relative positional embeddings like Transformer XL  
                \item \textbf{Application:} General language tasks including machine translation, question answering, abstractive summarization, and text classification
                \item \textbf{Date (of first known publication):} 10/2019
                \item \textbf{Num. Params:} 11 B (up to)
                \item \textbf{Corpus:} Colossal Clean Crawled Corpus (C4) — Cleaned up version of the Common Crawl dataset — 750 GB
                \item \textbf{License:} Open, Apache-2.0
                \item \textbf{Lab:} Google
            \end{itemize}
 
\subsection{Trajectory Transformers}

            \begin{itemize}
                \item \textbf{Reference:} \href{https://arxiv.org/abs/2106.02039}{\shortcite{janner2021offline}}
                \item \textbf{Link:} \url{https://trajectory-transformer.github.io}
                \item \textbf{Family:} GPT, Control Transformers” (not per se a family, but grouping here those Transformers that try to model more general control, RL-like, tasks) 
                \item \textbf{Pretraining Architecture:} Decoder
                \item \textbf{Pretraining Task:} predict most likely sequence
                \item \textbf{Extension:} Similarly to the Decision Transformers, the main extension introduced by Trajectory Transformers is a way to encode a trajectory (state, actions, rewards)  
                \item \textbf{Application:} General RL (reinforcement learning tasks)
                \item \textbf{Date (of first known publication):} 06/2021
                \item \textbf{Num. Params:} Smaller architecture than GPT
                \item \textbf{Corpus:} D4RL dataset and other RL datasets depending on the task at hand
                \item \textbf{License:} Open, MIT license
                \item \textbf{Lab:} UC Berkeley
            \end{itemize}

\subsection{Transformer XL}

            \begin{itemize}
                \item \textbf{Reference:} \shortcite{dai2019transformer}
                \item \textbf{Link:} \url{https://huggingface.co/docs/transformers/model_doc/transfo-xl}
                \item \textbf{Family:} 
                \item \textbf{Pretraining Architecture:} Decoder
                \item \textbf{Pretraining Task:} LM
                \item \textbf{Extension:} Relative positioned embeddings enable longer-context attention when compared to vanilla Transformer model  
                \item \textbf{Application:} General language tasks
                \item \textbf{Date (of first known publication):} 01/2019
                \item \textbf{Num. Params:} 151M
                \item \textbf{Corpus:} Different training datasets depending on experiments, but baseline is Wikitext-103
                 \item \textbf{License:} N/A
                \item \textbf{Lab:} CMU/Google
            \end{itemize}

\subsection{Turing-NLG}

            \begin{itemize}
                \item \textbf{Reference:} \href{https://www.microsoft.com/en-us/research/blog/turing-nlg-a-17-billion-parameter-language-model-by-microsoft}{\shortcite{rosset2020turing}}
                \item \textbf{Link:} N/A
                \item \textbf{Family:} GPT
                \item \textbf{Pretraining Architecture:} Decoder
                \item \textbf{Pretraining Task:} LM
                \item \textbf{Extension:} Optimized version of GPT2 with optimal hyperparameters and software/hardware platform to improve training
                \item \textbf{Application:} Same as GPT-2/3
                \item \textbf{Date (of first known publication):} 02/2020
                \item \textbf{Num. Params:} 17B originally, up to 530B more recently
                \item \textbf{Corpus:} Highest quality subset from The Pile + 2 CC snapshots (339B tokens)
                \item \textbf{License:} N/A
                \item \textbf{Lab:} Microsoft
            \end{itemize}

\subsection{UL2}

            \begin{itemize}
                \item \textbf{Reference:} \href{https://arxiv.org/abs/2205.05131}{\shortcite{yi2022ul2}}
                \item \textbf{Link:} \url{https://github.com/google-research/google-research/tree/master/ul2}
                \item \textbf{Family:} Transformer
                \item \textbf{Pretraining Architecture:} Encoder/Decoder
                \item \textbf{Pretraining Task:} Mixture-of-Denoisers, which combines diverse pre-training paradigms together
                \item \textbf{Extension:} UL2-20B (Unifying Language Learning) can be interpreted as a model that is quite similar to T5 but trained with a different objective and slightly different scaling knobs.
                \item \textbf{Application:} A unified framework for pre-training models that are universally effective across datasets and setups.
                \item \textbf{Date (of first known publication):} 05/2022
                \item \textbf{Num. Params:} 20B
                \item \textbf{Corpus:} 1 trillion tokens on C4
                \item \textbf{License:} Open, Apache-2.0
                \item \textbf{Lab:} Google
            \end{itemize}

\subsection{Vicuna}
            \begin{itemize}
                \item \textbf{Reference:} N/A
                \item \textbf{Link:} \url{https://vicuna.lmsys.org}
                \item \textbf{Family:} LLaMA
                \item \textbf{Pretraining Architecture:} Decoder
                \item \textbf{Fine-tuning Task:} human instructions
                \item \textbf{Extension:} LLaMA fine-tuned on user-shared conversations collected from ShareGPT.
                \item \textbf{Application:} Same as ChatGPT
                \item \textbf{Date (of first known publication):} 03/2023
                \item \textbf{Num. Params:} 13B
                \item \textbf{Corpus:} Conversations collected from ShareGPT
                \item \textbf{License:} Limited, Non-commercial bespoke license
                \item \textbf{Lab:} UC Berkeley, CMU, Stanford, UC San Diego, and MBZUAI
            \end{itemize}
            
\subsection{ViT}

            \begin{itemize}
                 \item \textbf{Reference:} \shortcite{dosovitskiy2020image}
                 \item \textbf{Link:} \url{https://huggingface.co/docs/transformers/model_doc/vit}
                 \item \textbf{Family:} BERT
                \item \textbf{Pretraining Architecture:} Encoder
                \item \textbf{Pretraining Task:} Image classification
                \item \textbf{Extension:} Extension of BERT architecture to train on patches of images
                \item \textbf{Application:} Image classification
                \item \textbf{Date (of first known publication):} 10/2020
                \item \textbf{Num. Params:} 86M(Base) to 632M (Huge)
                \item \textbf{Corpus:} From standard Imagenet to JFT-300M (large inhouse dataset)
                \item \textbf{License:} N/A
                \item \textbf{Lab:} Google
            \end{itemize}

\subsection{Wu Dao 2.0}

            \begin{itemize}
                \item \textbf{Reference:} See Wikipedia page\footnote{\url{https://en.wikipedia.org/wiki/Wu_Dao}}
                \item \textbf{Link:} See blog post\footnote{\url{https://mp.weixin.qq.com/s/BUQWZ5EdR19i40GuFofpBg}}
                \item \textbf{Family:} GLM (General Language Model)
                \item \textbf{Pretraining Architecture:} Decoder
                \item \textbf{Pretraining Task:} Autoregressive blank infilling
                \item \textbf{Extension:} Similar to GPT in that it uses a Decoder/autoregressive architecture but applies a different pretraining task proposed in the GLM family of models. Besides, Wu Dao uses a Fast Mixture of Experts (see \url{https://github.com/laekov/fastmoe)} approach to scale training to trillions of parameters
                \item \textbf{Application:} Language and multimodal (particularly image)
                \item \textbf{Date (of first known publication):} 06/2021
                \item \textbf{Num. Params:} 1.75T
                \item \textbf{Corpus:}  4.9 TB of high quality images and texts in both English and Chinese
                \item \textbf{License:} Closed source
                \item \textbf{Lab:} Beijing Academy of Artificial Intelligence
            \end{itemize}

\subsection{XLM-RoBERTa}

            \begin{itemize}
                \item \textbf{Reference:} \shortcite{conneau2019unsupervised}
                \item \textbf{Link:} \url{https://huggingface.co/docs/transformers/model_doc/xlm-roberta}
                \item \textbf{Family:} RoBERTa
                \item \textbf{Pretraining Architecture:} Encoder
                \item \textbf{Pretraining Task:} MLM (Dynamic)
                \item \textbf{Extension:} An extension of RoBERTa that introduces parameter tuning insights in the context of multilingual applications
                \item \textbf{Application:} Translation and other cross-lingual language tasks
                \item \textbf{Date (of first known publication):} 10/2019
                \item \textbf{Num. Params:} Base = 270M, Large = 550M
                \item \textbf{Corpus:} Cleaned Common Crawl in 100 languages
                \item \textbf{License:} Open, MIT license
                \item \textbf{Lab:} Facebook
            \end{itemize}

\subsection{XLNet}

            \begin{itemize}
                \item \textbf{Reference:} \shortcite{yang2019xlnet}
                \item \textbf{Link:} \url{https://huggingface.co/docs/transformers/model_doc/xlnet}
                \item \textbf{Family:} Transformer XL
                \item \textbf{Pretraining Architecture:} Decoder
                \item \textbf{Pretraining Task:} PLM
                \item \textbf{Extension:} This model basically adapts Transformer XL architecture to permutation-based LM
                \item \textbf{Application:} General language tasks
                \item \textbf{Date (of first known publication):} 05/2019
                \item \textbf{Num. Params:} Base=117M, Large=360M
                \item \textbf{Corpus:} Same as BERT + Giga5 (16GB text) + and aggressively filtered ClueWeb 2012-B (19GB), Common Crawl (110 GB)
                \item \textbf{License:} Open, MIT license
                \item \textbf{Lab:} CMU/Google
            \end{itemize}

\bibliography{references} 

\begin{thebibliography}{}

\bibitem[\protect\BCAY{Aghajanyan, Gupta, Shrivastava, Chen, Zettlemoyer, \BBA\
  Gupta}{Aghajanyan et~al.}{2021}]{aghajanyan2021muppet}
Aghajanyan, A., Gupta, A., Shrivastava, A., Chen, X., Zettlemoyer, L., \BBA\
  Gupta, S. \BBOP2021\BBCP.
\newblock \BBOQ Muppet: Massive Multi-task Representations with
  Pre-Finetuning\BBCQ\
\newblock \url{https://arxiv.org/abs/2101.11038}.

\bibitem[\protect\BCAY{Aghajanyan, Huang, Ross, Karpukhin, Xu, Goyal, Okhonko,
  Joshi, Ghosh, Lewis, et~al.}{Aghajanyan et~al.}{2022}]{aghajanyan2022cm3}
Aghajanyan, A., Huang, B., Ross, C., Karpukhin, V., Xu, H., Goyal, N., Okhonko,
  D., Joshi, M., Ghosh, G., Lewis, M., et~al. \BBOP2022\BBCP.
\newblock \BBOQ Cm3: A causal masked multimodal model of the internet\BBCQ\
\newblock \url{https://arxiv.org/abs/2201.07520}.

\bibitem[\protect\BCAY{Aghajanyan, Okhonko, Lewis, Joshi, Xu, Ghosh, \BBA\
  Zettlemoyer}{Aghajanyan et~al.}{2021}]{aghajanyan2021htlm}
Aghajanyan, A., Okhonko, D., Lewis, M., Joshi, M., Xu, H., Ghosh, G., \BBA\
  Zettlemoyer, L. \BBOP2021\BBCP.
\newblock \BBOQ Htlm: Hyper-text pre-training and prompting of language
  models\BBCQ\
\newblock \url{https://arxiv.org/abs/2107.06955}.

\bibitem[\protect\BCAY{Alayrac, Donahue, Luc, Miech, Barr, Hasson, Lenc,
  Mensch, Millican, Reynolds, et~al.}{Alayrac
  et~al.}{2022}]{alayrac2022flamingo}
Alayrac, J.-B., Donahue, J., Luc, P., Miech, A., Barr, I., Hasson, Y., Lenc,
  K., Mensch, A., Millican, K., Reynolds, M., et~al. \BBOP2022\BBCP.
\newblock \BBOQ Flamingo: a visual language model for few-shot learning\BBCQ\
\newblock \url{https://arxiv.org/abs/2204.14198}.

\bibitem[\protect\BCAY{Askell, Bai, Chen, Drain, Ganguli, Henighan, Jones,
  Joseph, Mann, DasSarma, et~al.}{Askell et~al.}{2021}]{askell2021general}
Askell, A., Bai, Y., Chen, A., Drain, D., Ganguli, D., Henighan, T., Jones, A.,
  Joseph, N., Mann, B., DasSarma, N., et~al. \BBOP2021\BBCP.
\newblock \BBOQ A general language assistant as a laboratory for
  alignment\BBCQ\
\newblock \url{https://arxiv.org/abs/2112.00861}.

\bibitem[\protect\BCAY{Bai, Jones, Ndousse, Askell, Chen, DasSarma, Drain,
  Fort, Ganguli, Henighan, Joseph, Kadavath, Kernion, Conerly, El-Showk,
  Elhage, Hatfield-Dodds, Hernandez, Hume, Johnston, Kravec, Lovitt, Nanda,
  Olsson, Amodei, Brown, Clark, McCandlish, Olah, Mann, \BBA\ Kaplan}{Bai
  et~al.}{2022a}]{bai2022training}
Bai, Y., Jones, A., Ndousse, K., Askell, A., Chen, A., DasSarma, N., Drain, D.,
  Fort, S., Ganguli, D., Henighan, T., Joseph, N., Kadavath, S., Kernion, J.,
  Conerly, T., El-Showk, S., Elhage, N., Hatfield-Dodds, Z., Hernandez, D.,
  Hume, T., Johnston, S., Kravec, S., Lovitt, L., Nanda, N., Olsson, C.,
  Amodei, D., Brown, T., Clark, J., McCandlish, S., Olah, C., Mann, B., \BBA\
  Kaplan, J. \BBOP2022a\BBCP.
\newblock \BBOQ Training a helpful and harmless assistant with reinforcement
  learning from human feedback\BBCQ\
\newblock \url{https://arxiv.org/abs/2204.05862}.

\bibitem[\protect\BCAY{Bai, Kadavath, Kundu, Askell, Kernion, Jones, Chen,
  Goldie, Mirhoseini, McKinnon, Chen, Olsson, Olah, Hernandez, Drain, Ganguli,
  Li, Tran-Johnson, Perez, Kerr, Mueller, Ladish, Landau, Ndousse, Lukosuite,
  Lovitt, Sellitto, Elhage, Schiefer, Mercado, DasSarma, Lasenby, Larson,
  Ringer, Johnston, Kravec, Showk, Fort, Lanham, Telleen-Lawton, Conerly,
  Henighan, Hume, Bowman, Hatfield-Dodds, Mann, Amodei, Joseph, McCandlish,
  Brown, \BBA\ Kaplan}{Bai et~al.}{2022b}]{bai2022constitutional}
Bai, Y., Kadavath, S., Kundu, S., Askell, A., Kernion, J., Jones, A., Chen, A.,
  Goldie, A., Mirhoseini, A., McKinnon, C., Chen, C., Olsson, C., Olah, C.,
  Hernandez, D., Drain, D., Ganguli, D., Li, D., Tran-Johnson, E., Perez, E.,
  Kerr, J., Mueller, J., Ladish, J., Landau, J., Ndousse, K., Lukosuite, K.,
  Lovitt, L., Sellitto, M., Elhage, N., Schiefer, N., Mercado, N., DasSarma,
  N., Lasenby, R., Larson, R., Ringer, S., Johnston, S., Kravec, S., Showk,
  S.~E., Fort, S., Lanham, T., Telleen-Lawton, T., Conerly, T., Henighan, T.,
  Hume, T., Bowman, S.~R., Hatfield-Dodds, Z., Mann, B., Amodei, D., Joseph,
  N., McCandlish, S., Brown, T., \BBA\ Kaplan, J. \BBOP2022b\BBCP.
\newblock \BBOQ Constitutional AI: Harmlessness from AI Feedback\BBCQ\
\newblock \url{https://arxiv.org/abs/2212.08073}.

\bibitem[\protect\BCAY{Baidoo-Anu \BBA\ Owusu~Ansah}{Baidoo-Anu \BBA\
  Owusu~Ansah}{2023}]{baidoo2023education}
Baidoo-Anu, D.\BBACOMMA\  \BBA\ Owusu~Ansah, L. \BBOP2023\BBCP.
\newblock \BBOQ Education in the era of generative artificial intelligence
  (AI): Understanding the potential benefits of ChatGPT in promoting teaching
  and learning\BBCQ\
\newblock \url{https://papers.ssrn.com/sol3/papers.cfm?abstract_id=4337484}.

\bibitem[\protect\BCAY{Biderman, Schoelkopf, \BBA\ and}{Biderman
  et~al.}{2023}]{biderman2023pythia}
Biderman, S., Schoelkopf, H., \BBA\ and, Q.~A. \BBOP2023\BBCP.
\newblock \BBOQ Pythia: A Suite for Analyzing Large Language Models Across
  Training and Scaling\BBCQ\
\newblock \url{https://arxiv.org/abs/2304.01373}.

\bibitem[\protect\BCAY{Black, Biderman, Hallahan, Anthony, Gao, Golding, He,
  Leahy, McDonell, Phang, et~al.}{Black et~al.}{2022}]{black2022gpt}
Black, S., Biderman, S., Hallahan, E., Anthony, Q., Gao, L., Golding, L., He,
  H., Leahy, C., McDonell, K., Phang, J., et~al. \BBOP2022\BBCP.
\newblock \BBOQ Gpt-neox-20b: An open-source autoregressive language
  model\BBCQ\
\newblock \url{https://arxiv.org/abs/2204.06745}.

\bibitem[\protect\BCAY{Bommasani, Hudson, Adeli, Altman, Arora, von Arx,
  Bernstein, Bohg, Bosselut, Brunskill, et~al.}{Bommasani
  et~al.}{2021}]{bommasani2021opportunities}
Bommasani, R., Hudson, D.~A., Adeli, E., Altman, R., Arora, S., von Arx, S.,
  Bernstein, M.~S., Bohg, J., Bosselut, A., Brunskill, E., et~al.
  \BBOP2021\BBCP.
\newblock \BBOQ On the opportunities and risks of foundation models\BBCQ\
\newblock \url{https://arxiv.org/abs/2108.07258}.

\bibitem[\protect\BCAY{Brown, Mann, Ryder, Subbiah, Kaplan, Dhariwal,
  Neelakantan, Shyam, Sastry, Askell, et~al.}{Brown
  et~al.}{2020}]{brown2020language}
Brown, T., Mann, B., Ryder, N., Subbiah, M., Kaplan, J.~D., Dhariwal, P.,
  Neelakantan, A., Shyam, P., Sastry, G., Askell, A., et~al. \BBOP2020\BBCP.
\newblock \BBOQ Language models are few-shot learners\BBCQ\
\newblock {\Bem Advances in Neural Information Processing Systems}, {\Bem 33},
  1877--1901.

\bibitem[\protect\BCAY{Chen, Lu, Rajeswaran, Lee, Grover, Laskin, Abbeel,
  Srinivas, \BBA\ Mordatch}{Chen et~al.}{2021}]{chen2021decision}
Chen, L., Lu, K., Rajeswaran, A., Lee, K., Grover, A., Laskin, M., Abbeel, P.,
  Srinivas, A., \BBA\ Mordatch, I. \BBOP2021\BBCP.
\newblock \BBOQ Decision transformer: Reinforcement learning via sequence
  modeling\BBCQ\
\newblock {\Bem Advances in Neural Information Processing Systems}, {\Bem 34},
  15084--15097.

\bibitem[\protect\BCAY{Cho, Van~Merri{\"e}nboer, Bahdanau, \BBA\ Bengio}{Cho
  et~al.}{2014}]{cho2014properties}
Cho, K., Van~Merri{\"e}nboer, B., Bahdanau, D., \BBA\ Bengio, Y.
  \BBOP2014\BBCP.
\newblock \BBOQ On the properties of neural machine translation:
  Encoder-decoder approaches\BBCQ\
\newblock \url{https://arxiv.org/abs/1409.1259}.

\bibitem[\protect\BCAY{Chowdhery, Narang, Devlin, Bosma, Mishra, Roberts,
  Barham, Chung, Sutton, Gehrmann, et~al.}{Chowdhery
  et~al.}{2022}]{chowdhery2022palm}
Chowdhery, A., Narang, S., Devlin, J., Bosma, M., Mishra, G., Roberts, A.,
  Barham, P., Chung, H.~W., Sutton, C., Gehrmann, S., et~al. \BBOP2022\BBCP.
\newblock \BBOQ Palm: Scaling language modeling with pathways\BBCQ\
\newblock \url{https://arxiv.org/abs/2204.02311}.

\bibitem[\protect\BCAY{Christiano, Leike, Brown, Martic, Legg, \BBA\
  Amodei}{Christiano et~al.}{2023}]{christiano2023deep}
Christiano, P., Leike, J., Brown, T.~B., Martic, M., Legg, S., \BBA\ Amodei, D.
  \BBOP2023\BBCP.
\newblock \BBOQ Deep reinforcement learning from human preferences\BBCQ\
\newblock \url{https://arxiv.org/abs/1706.03741}.

\bibitem[\protect\BCAY{Chung, Hou, Longpre, Zoph, Tay, Fedus, Li, Wang,
  Dehghani, Brahma, Webson, Gu, Dai, Suzgun, Chen, Chowdhery, Castro-Ros,
  Pellat, Robinson, Valter, Narang, Mishra, Yu, Zhao, Huang, Dai, Yu, Petrov,
  Chi, Dean, Devlin, Roberts, Zhou, Le, \BBA\ Wei}{Chung
  et~al.}{2022}]{hyung2022flant5}
Chung, H.~W., Hou, L., Longpre, S., Zoph, B., Tay, Y., Fedus, W., Li, Y., Wang,
  X., Dehghani, M., Brahma, S., Webson, A., Gu, S.~S., Dai, Z., Suzgun, M.,
  Chen, X., Chowdhery, A., Castro-Ros, A., Pellat, M., Robinson, K., Valter,
  D., Narang, S., Mishra, G., Yu, A., Zhao, V., Huang, Y., Dai, A., Yu, H.,
  Petrov, S., Chi, E.~H., Dean, J., Devlin, J., Roberts, A., Zhou, D., Le,
  Q.~V., \BBA\ Wei, J. \BBOP2022\BBCP.
\newblock \BBOQ Scaling Instruction-Finetuned Language Models\BBCQ\
\newblock \url{https://arxiv.org/abs/2210.11416}.

\bibitem[\protect\BCAY{Clark, Luong, Le, \BBA\ Manning}{Clark
  et~al.}{2020}]{clark2020electra}
Clark, K., Luong, M.-T., Le, Q.~V., \BBA\ Manning, C.~D. \BBOP2020\BBCP.
\newblock \BBOQ Electra: Pre-training text encoders as discriminators rather
  than generators\BBCQ\
\newblock \url{https://arxiv.org/abs/2003.10555}.

\bibitem[\protect\BCAY{Conneau, Khandelwal, Goyal, Chaudhary, Wenzek,
  Guzm{\'a}n, Grave, Ott, Zettlemoyer, \BBA\ Stoyanov}{Conneau
  et~al.}{2019}]{conneau2019unsupervised}
Conneau, A., Khandelwal, K., Goyal, N., Chaudhary, V., Wenzek, G., Guzm{\'a}n,
  F., Grave, E., Ott, M., Zettlemoyer, L., \BBA\ Stoyanov, V. \BBOP2019\BBCP.
\newblock \BBOQ Unsupervised cross-lingual representation learning at
  scale\BBCQ\
\newblock \url{https://arxiv.org/abs/1911.02116}.

\bibitem[\protect\BCAY{Dai, Yang, Yang, Carbonell, Le, \BBA\ Salakhutdinov}{Dai
  et~al.}{2019}]{dai2019transformer}
Dai, Z., Yang, Z., Yang, Y., Carbonell, J., Le, Q.~V., \BBA\ Salakhutdinov, R.
  \BBOP2019\BBCP.
\newblock \BBOQ Transformer-xl: Attentive language models beyond a fixed-length
  context\BBCQ\
\newblock \url{https://arxiv.org/abs/1901.02860}.

\bibitem[\protect\BCAY{Dennean, Gantori, Limas, \BBA\ Allen~Pu}{Dennean
  et~al.}{2023}]{UBSChatGPT2023}
Dennean, K., Gantori, S., Limas, D.~K., \BBA\ Allen~Pu, a. R.~G.
  \BBOP2023\BBCP.
\newblock \BBOQ Let's chat about ChatGPT\BBCQ\
\newblock
  \url{https://www.ubs.com/global/en/wealth-management/our-approach/marketnews/article.1585717.html}.

\bibitem[\protect\BCAY{Devlin, Chang, Lee, \BBA\ Toutanova}{Devlin
  et~al.}{2018}]{devlin2018bert}
Devlin, J., Chang, M.-W., Lee, K., \BBA\ Toutanova, K. \BBOP2018\BBCP.
\newblock \BBOQ Bert: Pre-training of deep bidirectional transformers for
  language understanding\BBCQ\
\newblock \url{https://arxiv.org/abs/1810.04805}.

\bibitem[\protect\BCAY{Dosovitskiy, Beyer, Kolesnikov, Weissenborn, Zhai,
  Unterthiner, Dehghani, Minderer, Heigold, Gelly, et~al.}{Dosovitskiy
  et~al.}{2020}]{dosovitskiy2020image}
Dosovitskiy, A., Beyer, L., Kolesnikov, A., Weissenborn, D., Zhai, X.,
  Unterthiner, T., Dehghani, M., Minderer, M., Heigold, G., Gelly, S., et~al.
  \BBOP2020\BBCP.
\newblock \BBOQ An image is worth 16x16 words: Transformers for image
  recognition at scale\BBCQ\
\newblock \url{https://arxiv.org/abs/2010.11929}.

\bibitem[\protect\BCAY{Du, Huang, Dai, Tong, Lepikhin, Xu, Krikun, Zhou, Yu,
  Firat, et~al.}{Du et~al.}{2022a}]{du2022glam}
Du, N., Huang, Y., Dai, A.~M., Tong, S., Lepikhin, D., Xu, Y., Krikun, M.,
  Zhou, Y., Yu, A.~W., Firat, O., et~al. \BBOP2022a\BBCP.
\newblock \BBOQ Glam: Efficient scaling of language models with
  mixture-of-experts\BBCQ\
\newblock In {\Bem International Conference on Machine Learning}, \BPGS\
  5547--5569. PMLR.

\bibitem[\protect\BCAY{Du, Qian, Liu, Ding, Qiu, Yang, \BBA\ Tang}{Du
  et~al.}{2022b}]{zhengxiao2022glm}
Du, Z., Qian, Y., Liu, X., Ding, M., Qiu, J., Yang, Z., \BBA\ Tang, J.
  \BBOP2022b\BBCP.
\newblock \BBOQ GLM: General Language Model Pretraining with Autoregressive
  Blank Infilling\BBCQ\
\newblock \url{https://arxiv.org/abs/2103.10360}.

\bibitem[\protect\BCAY{Esser, Rombach, \BBA\ Ommer}{Esser
  et~al.}{2021}]{esser2021taming}
Esser, P., Rombach, R., \BBA\ Ommer, B. \BBOP2021\BBCP.
\newblock \BBOQ Taming transformers for high-resolution image synthesis\BBCQ\
\newblock In {\Bem Proceedings of the IEEE/CVF conference on computer vision
  and pattern recognition}, \BPGS\ 12873--12883.

\bibitem[\protect\BCAY{Fedus, Zoph, \BBA\ Shazeer}{Fedus
  et~al.}{2021}]{fedus2021switch}
Fedus, W., Zoph, B., \BBA\ Shazeer, N. \BBOP2021\BBCP.
\newblock \BBOQ Switch transformers: Scaling to trillion parameter models with
  simple and efficient sparsity\BBCQ\
\newblock {\Bem Journal of Machine Learning Research}, {\Bem 23}, 1--40.

\bibitem[\protect\BCAY{Fuchs, Worrall, Fischer, \BBA\ Welling}{Fuchs
  et~al.}{2020}]{fuchs2020se3}
Fuchs, F.~B., Worrall, D.~E., Fischer, V., \BBA\ Welling, M. \BBOP2020\BBCP.
\newblock \BBOQ SE(3)-Transformers: 3D roto-translation equivariant attention
  networks\BBCQ\
\newblock \url{https://arxiv.org/abs/2006.10503}.

\bibitem[\protect\BCAY{Glaese, McAleese, Tr{\k{e}}bacz, Aslanides, Firoiu,
  Ewalds, Rauh, Weidinger, Chadwick, Thacker, et~al.}{Glaese
  et~al.}{2022}]{glaese2022improving}
Glaese, A., McAleese, N., Tr{\k{e}}bacz, M., Aslanides, J., Firoiu, V., Ewalds,
  T., Rauh, M., Weidinger, L., Chadwick, M., Thacker, P., et~al.
  \BBOP2022\BBCP.
\newblock \BBOQ Improving alignment of dialogue agents via targeted human
  judgements\BBCQ\
\newblock \url{https://arxiv.org/abs/2209.14375}.

\bibitem[\protect\BCAY{Hatamizadeh, Yin, Kautz, \BBA\ Molchanov}{Hatamizadeh
  et~al.}{2022}]{hatamizadeh2022global}
Hatamizadeh, A., Yin, H., Kautz, J., \BBA\ Molchanov, P. \BBOP2022\BBCP.
\newblock \BBOQ Global context vision transformers\BBCQ\
\newblock \url{https://arxiv.org/abs/2206.09959}.

\bibitem[\protect\BCAY{He, Liu, Gao, \BBA\ Chen}{He
  et~al.}{2021}]{he2021deberta}
He, P., Liu, X., Gao, J., \BBA\ Chen, W. \BBOP2021\BBCP.
\newblock \BBOQ {DeBERTa: Decoding-enhanced BERT with disentangled
  attention}\BBCQ\
\newblock In {\Bem International Conference on Learning Representations}.

\bibitem[\protect\BCAY{Hochreiter \BBA\ Schmidhuber}{Hochreiter \BBA\
  Schmidhuber}{1997}]{hochreiter1997long}
Hochreiter, S.\BBACOMMA\  \BBA\ Schmidhuber, J. \BBOP1997\BBCP.
\newblock \BBOQ Long short-term memory\BBCQ\
\newblock {\Bem Neural computation}, {\Bem 9\/}(8), 1735--1780.

\bibitem[\protect\BCAY{Hoffmann, Borgeaud, Mensch, Buchatskaya, Cai,
  Rutherford, Casas, Hendricks, Welbl, Clark, et~al.}{Hoffmann
  et~al.}{2022}]{hoffmann2022training}
Hoffmann, J., Borgeaud, S., Mensch, A., Buchatskaya, E., Cai, T., Rutherford,
  E., Casas, D. d.~L., Hendricks, L.~A., Welbl, J., Clark, A., et~al.
  \BBOP2022\BBCP.
\newblock \BBOQ Training compute-optimal large language models\BBCQ\
\newblock \url{https://arxiv.org/abs/2203.15556}.

\bibitem[\protect\BCAY{Janner, Li, \BBA\ Levine}{Janner
  et~al.}{2021}]{janner2021offline}
Janner, M., Li, Q., \BBA\ Levine, S. \BBOP2021\BBCP.
\newblock \BBOQ Offline reinforcement learning as one big sequence modeling
  problem\BBCQ\
\newblock {\Bem Advances in Neural Information Processing Systems}, {\Bem 34},
  1273--1286.

\bibitem[\protect\BCAY{Jumper, Evans, Pritzel, Green, Figurnov, Ronneberger,
  Tunyasuvunakool, Bates, Zidek, Potapenko, et~al.}{Jumper
  et~al.}{2021}]{jumper2021highly}
Jumper, J., Evans, R., Pritzel, A., Green, T., Figurnov, M., Ronneberger, O.,
  Tunyasuvunakool, K., Bates, R., Zidek, A., Potapenko, A., et~al.
  \BBOP2021\BBCP.
\newblock \BBOQ Highly accurate protein structure prediction with
  AlphaFold\BBCQ\
\newblock {\Bem Nature}, {\Bem 596\/}(7873), 583--589.

\bibitem[\protect\BCAY{Keskar, McCann, Varshney, Xiong, \BBA\ Socher}{Keskar
  et~al.}{2019}]{keskar2019ctrl}
Keskar, N.~S., McCann, B., Varshney, L.~R., Xiong, C., \BBA\ Socher, R.
  \BBOP2019\BBCP.
\newblock \BBOQ Ctrl: A conditional transformer language model for controllable
  generation\BBCQ\
\newblock \url{https://arxiv.org/abs/1909.05858}.

\bibitem[\protect\BCAY{Köpf, Kilcher, von Rütte, Anagnostidis, Tam, Stevens,
  Barhoum, Duc, Stanley, Nagyfi, ES, Suri, Glushkov, Dantuluri, Maguire,
  Schuhmann, Nguyen, \BBA\ Mattick}{Köpf et~al.}{2023}]{kopf2022openassistant}
Köpf, A., Kilcher, Y., von Rütte, D., Anagnostidis, S., Tam, Z.-R., Stevens,
  K., Barhoum, A., Duc, N.~M., Stanley, O., Nagyfi, R., ES, S., Suri, S.,
  Glushkov, D., Dantuluri, A., Maguire, A., Schuhmann, C., Nguyen, H., \BBA\
  Mattick, A. \BBOP2023\BBCP.
\newblock \BBOQ OpenAssistant Conversations -- Democratizing Large Language
  Model Alignment\BBCQ\
\newblock \url{https://arxiv.org/abs/2304.07327}.

\bibitem[\protect\BCAY{Lan, Chen, Goodman, Gimpel, Sharma, \BBA\ Soricut}{Lan
  et~al.}{2019}]{lan2019albert}
Lan, Z., Chen, M., Goodman, S., Gimpel, K., Sharma, P., \BBA\ Soricut, R.
  \BBOP2019\BBCP.
\newblock \BBOQ ALBERT: A lite BERT for self-supervised learning of language
  representations\BBCQ\
\newblock \url{https://arxiv.org/abs/1909.11942}.

\bibitem[\protect\BCAY{Lewis, Liu, Goyal, Ghazvininejad, Mohamed, Levy,
  Stoyanov, \BBA\ Zettlemoyer}{Lewis et~al.}{2019}]{lewis2019bart}
Lewis, M., Liu, Y., Goyal, N., Ghazvininejad, M., Mohamed, A., Levy, O.,
  Stoyanov, V., \BBA\ Zettlemoyer, L. \BBOP2019\BBCP.
\newblock \BBOQ Bart: Denoising sequence-to-sequence pre-training for natural
  language generation, translation, and comprehension\BBCQ\
\newblock \url{https://arxiv.org/abs/1910.13461}.

\bibitem[\protect\BCAY{Lewkowycz, Andreassen, Dohan, Dyer, Michalewski,
  Ramasesh, Slone, Anil, Schlag, Gutman-Solo, et~al.}{Lewkowycz
  et~al.}{2022}]{lewkowycz2022solving}
Lewkowycz, A., Andreassen, A., Dohan, D., Dyer, E., Michalewski, H., Ramasesh,
  V., Slone, A., Anil, C., Schlag, I., Gutman-Solo, T., et~al. \BBOP2022\BBCP.
\newblock \BBOQ Solving quantitative reasoning problems with language
  models\BBCQ\
\newblock \url{https://arxiv.org/abs/2206.14858}.

\bibitem[\protect\BCAY{Li, Wang, Tan, Nallapati, Bhatia, Arnold, Xiang, \BBA\
  Roth}{Li et~al.}{2022}]{li2022dq}
Li, Z., Wang, Z., Tan, M., Nallapati, R., Bhatia, P., Arnold, A., Xiang, B.,
  \BBA\ Roth, D. \BBOP2022\BBCP.
\newblock \BBOQ DQ-BART: Efficient Sequence-to-Sequence Model via Joint
  Distillation and Quantization\BBCQ\
\newblock \url{https://arxiv.org/abs/2203.11239}.

\bibitem[\protect\BCAY{Lieber, Sharir, Lenz, \BBA\ Shoham}{Lieber
  et~al.}{2021}]{lieber2021jurassic}
Lieber, O., Sharir, O., Lenz, B., \BBA\ Shoham, Y. \BBOP2021\BBCP.
\newblock \BBOQ Jurassic-1: Technical details and evaluation\BBCQ\
\newblock
  \url{https://uploads-ssl.webflow.com/60fd4503684b466578c0d307/61138924626a6981ee09caf6_jurassic_tech_paper.pdf}.

\bibitem[\protect\BCAY{Liu, He, Chen, \BBA\ Gao}{Liu
  et~al.}{2019}]{liu2019mtdnn}
Liu, X., He, P., Chen, W., \BBA\ Gao, J. \BBOP2019\BBCP.
\newblock \BBOQ Multi-Task Deep Neural Networks for Natural Language
  Understanding\BBCQ\
\newblock In {\Bem Proceedings of the 57th Annual Meeting of the Association
  for Computational Linguistics}, \BPGS\ 4487--4496\ Florence, Italy.
  Association for Computational Linguistics.

\bibitem[\protect\BCAY{Liu, Gu, Goyal, Li, Edunov, Ghazvininejad, Lewis, \BBA\
  Zettlemoyer}{Liu et~al.}{2020}]{liu2020multilingual}
Liu, Y., Gu, J., Goyal, N., Li, X., Edunov, S., Ghazvininejad, M., Lewis, M.,
  \BBA\ Zettlemoyer, L. \BBOP2020\BBCP.
\newblock \BBOQ Multilingual denoising pre-training for neural machine
  translation\BBCQ\
\newblock {\Bem Transactions of the Association for Computational Linguistics},
  {\Bem 8}, 726--742.

\bibitem[\protect\BCAY{Liu, Ott, Goyal, Du, Joshi, Chen, Levy, Lewis,
  Zettlemoyer, \BBA\ Stoyanov}{Liu et~al.}{2019}]{liu2019roberta}
Liu, Y., Ott, M., Goyal, N., Du, J., Joshi, M., Chen, D., Levy, O., Lewis, M.,
  Zettlemoyer, L., \BBA\ Stoyanov, V. \BBOP2019\BBCP.
\newblock \BBOQ Roberta: A robustly optimized bert pretraining approach\BBCQ\
\newblock \url{https://arxiv.org/abs/1907.11692}.

\bibitem[\protect\BCAY{Liu, Lin, Cao, Hu, Wei, Zhang, Lin, \BBA\ Guo}{Liu
  et~al.}{2021}]{liu2021swin}
Liu, Z., Lin, Y., Cao, Y., Hu, H., Wei, Y., Zhang, Z., Lin, S., \BBA\ Guo, B.
  \BBOP2021\BBCP.
\newblock \BBOQ Swin transformer: Hierarchical vision transformer using shifted
  windows\BBCQ\
\newblock In {\Bem Proceedings of the IEEE/CVF international conference on
  computer vision}, \BPGS\ 10012--10022.

\bibitem[\protect\BCAY{Menick, Trebacz, Mikulik, Aslanides, Song, Chadwick,
  Glaese, Young, Campbell-Gillingham, Irving, et~al.}{Menick
  et~al.}{2022}]{menick2022teaching}
Menick, J., Trebacz, M., Mikulik, V., Aslanides, J., Song, F., Chadwick, M.,
  Glaese, M., Young, S., Campbell-Gillingham, L., Irving, G., et~al.
  \BBOP2022\BBCP.
\newblock \BBOQ Teaching language models to support answers with verified
  quotes\BBCQ\
\newblock \url{https://arxiv.org/abs/2203.11147}.

\bibitem[\protect\BCAY{Mikolov, Karafiat, Burget, Cernocky, \BBA\
  Khudanpur}{Mikolov et~al.}{2010}]{mikolov2010recurrent}
Mikolov, T., Karafiat, M., Burget, L., Cernocky, J., \BBA\ Khudanpur, S.
  \BBOP2010\BBCP.
\newblock \BBOQ Recurrent neural network based language model\BBCQ\
\newblock In {\Bem Interspeech}, \lowercase{\BVOL}~2, \BPGS\ 1045--1048.

\bibitem[\protect\BCAY{Minaee, Mikolov, Nikzad, Chenaghlu, Socher, Amatriain,
  \BBA\ Gao}{Minaee et~al.}{2024}]{minaee2024large}
Minaee, S., Mikolov, T., Nikzad, N., Chenaghlu, M., Socher, R., Amatriain, X.,
  \BBA\ Gao, J. \BBOP2024\BBCP.
\newblock \BBOQ Large Language Models: A Survey\BBCQ.

\bibitem[\protect\BCAY{Nichol, Dhariwal, Ramesh, Shyam, Mishkin, McGrew,
  Sutskever, \BBA\ Chen}{Nichol et~al.}{2021}]{nichol2021glide}
Nichol, A., Dhariwal, P., Ramesh, A., Shyam, P., Mishkin, P., McGrew, B.,
  Sutskever, I., \BBA\ Chen, M. \BBOP2021\BBCP.
\newblock \BBOQ Glide: Towards photorealistic image generation and editing with
  text-guided diffusion models\BBCQ\
\newblock \url{https://arxiv.org/abs/2112.10741}.

\bibitem[\protect\BCAY{OpenAI}{OpenAI}{2023}]{gpt-4}
OpenAI \BBOP2023\BBCP.
\newblock \BBOQ GPT-4 Technical Report\BBCQ\
\newblock \url{https://arxiv.org/abs/2303.08774}.

\bibitem[\protect\BCAY{Ouyang, Wu, Jiang, Almeida, Wainwright, Mishkin, Zhang,
  Agarwal, Slama, Ray, et~al.}{Ouyang et~al.}{2022}]{ouyang2022training}
Ouyang, L., Wu, J., Jiang, X., Almeida, D., Wainwright, C.~L., Mishkin, P.,
  Zhang, C., Agarwal, S., Slama, K., Ray, A., et~al. \BBOP2022\BBCP.
\newblock \BBOQ Training language models to follow instructions with human
  feedback\BBCQ\
\newblock \url{https://arxiv.org/abs/2203.02155}.

\bibitem[\protect\BCAY{Qiu, Sun, Xu, Shao, Dai, \BBA\ Huang}{Qiu
  et~al.}{2020}]{qiu2020pre}
Qiu, X., Sun, T., Xu, Y., Shao, Y., Dai, N., \BBA\ Huang, X. \BBOP2020\BBCP.
\newblock \BBOQ Pre-trained models for natural language processing: A
  survey\BBCQ\
\newblock {\Bem Science China Technological Sciences}, {\Bem 63\/}(10),
  1872--1897.

\bibitem[\protect\BCAY{Radford, Kim, Hallacy, Ramesh, Goh, Agarwal, Sastry,
  Askell, Mishkin, Clark, et~al.}{Radford et~al.}{2021}]{radford2021learning}
Radford, A., Kim, J.~W., Hallacy, C., Ramesh, A., Goh, G., Agarwal, S., Sastry,
  G., Askell, A., Mishkin, P., Clark, J., et~al. \BBOP2021\BBCP.
\newblock \BBOQ Learning transferable visual models from natural language
  supervision\BBCQ\
\newblock In {\Bem International conference on machine learning}, \BPGS\
  8748--8763. PMLR.

\bibitem[\protect\BCAY{Radford, Narasimhan, Salimans, Sutskever,
  et~al.}{Radford et~al.}{2018}]{radford2018improving}
Radford, A., Narasimhan, K., Salimans, T., Sutskever, I., et~al.
  \BBOP2018\BBCP.
\newblock \BBOQ Improving language understanding by generative
  pre-training\BBCQ\
\newblock
  \url{https://cdn.openai.com/research-covers/language-unsupervised/language_understanding_paper.pdf}.

\bibitem[\protect\BCAY{Radford, Wu, Child, Luan, Amodei, Sutskever,
  et~al.}{Radford et~al.}{2019}]{radford2019language}
Radford, A., Wu, J., Child, R., Luan, D., Amodei, D., Sutskever, I., et~al.
  \BBOP2019\BBCP.
\newblock \BBOQ Language models are unsupervised multitask learners\BBCQ\
\newblock
  \url{https://paperswithcode.com/paper/language-models-are-unsupervised-multitask}.

\bibitem[\protect\BCAY{Rae, Borgeaud, Cai, Millican, Hoffmann, Song, Aslanides,
  Henderson, Ring, Young, et~al.}{Rae et~al.}{2021}]{rae2021scaling}
Rae, J.~W., Borgeaud, S., Cai, T., Millican, K., Hoffmann, J., Song, F.,
  Aslanides, J., Henderson, S., Ring, R., Young, S., et~al. \BBOP2021\BBCP.
\newblock \BBOQ Scaling language models: Methods, analysis \& insights from
  training gopher\BBCQ\
\newblock \url{https://arxiv.org/abs/2112.11446}.

\bibitem[\protect\BCAY{Raffel, Shazeer, Roberts, Lee, Narang, Matena, Zhou, Li,
  \BBA\ Liu}{Raffel et~al.}{2020}]{raffel2020exploring}
Raffel, C., Shazeer, N., Roberts, A., Lee, K., Narang, S., Matena, M., Zhou,
  Y., Li, W., \BBA\ Liu, P.~J. \BBOP2020\BBCP.
\newblock \BBOQ Exploring the limits of transfer learning with a unified
  text-to-text transformer\BBCQ\
\newblock {\Bem The Journal of Machine Learning Research}, {\Bem 21\/}(1),
  5485--5551.

\bibitem[\protect\BCAY{Ramesh, Dhariwal, Nichol, Chu, \BBA\ Chen}{Ramesh
  et~al.}{2022}]{ramesh2022hierarchical}
Ramesh, A., Dhariwal, P., Nichol, A., Chu, C., \BBA\ Chen, M. \BBOP2022\BBCP.
\newblock \BBOQ Hierarchical text-conditional image generation with clip
  latents\BBCQ\
\newblock \url{https://arxiv.org/abs/2204.06125}.

\bibitem[\protect\BCAY{Ramesh, Pavlov, Goh, Gray, Voss, Radford, Chen, \BBA\
  Sutskever}{Ramesh et~al.}{2021}]{ramesh2021zero}
Ramesh, A., Pavlov, M., Goh, G., Gray, S., Voss, C., Radford, A., Chen, M.,
  \BBA\ Sutskever, I. \BBOP2021\BBCP.
\newblock \BBOQ Zero-shot text-to-image generation\BBCQ\
\newblock In {\Bem International Conference on Machine Learning}, \BPGS\
  8821--8831. PMLR.

\bibitem[\protect\BCAY{Reed, Zolna, Parisotto, Colmenarejo, Novikov,
  Barth-Maron, Gimenez, Sulsky, Kay, Springenberg, et~al.}{Reed
  et~al.}{2022}]{reed2022generalist}
Reed, S., Zolna, K., Parisotto, E., Colmenarejo, S.~G., Novikov, A.,
  Barth-Maron, G., Gimenez, M., Sulsky, Y., Kay, J., Springenberg, J.~T.,
  et~al. \BBOP2022\BBCP.
\newblock \BBOQ A generalist agent\BBCQ\
\newblock \url{https://arxiv.org/abs/2205.06175}.

\bibitem[\protect\BCAY{Rombach, Blattmann, Lorenz, Esser, \BBA\ Ommer}{Rombach
  et~al.}{2022}]{rombach2022high}
Rombach, R., Blattmann, A., Lorenz, D., Esser, P., \BBA\ Ommer, B.
  \BBOP2022\BBCP.
\newblock \BBOQ High-resolution image synthesis with latent diffusion
  models\BBCQ\
\newblock In {\Bem Proceedings of the IEEE/CVF Conference on Computer Vision
  and Pattern Recognition}, \BPGS\ 10684--10695.

\bibitem[\protect\BCAY{Rosset}{Rosset}{2020}]{rosset2020turing}
Rosset, C. \BBOP2020\BBCP.
\newblock \BBOQ Turing-NLG: A 17-billion-parameter language model by
  Microsoft\BBCQ\
\newblock
  \url{https://www.microsoft.com/en-us/research/blog/turing-nlg-a-17-billion-parameter-language-model-by-microsoft/}.

\bibitem[\protect\BCAY{Saharia, Chan, Saxena, Li, Whang, Denton, Ghasemipour,
  Ayan, Mahdavi, Lopes, et~al.}{Saharia
  et~al.}{2022}]{saharia2022photorealistic}
Saharia, C., Chan, W., Saxena, S., Li, L., Whang, J., Denton, E., Ghasemipour,
  S. K.~S., Ayan, B.~K., Mahdavi, S.~S., Lopes, R.~G., et~al. \BBOP2022\BBCP.
\newblock \BBOQ Photorealistic text-to-image diffusion models with deep
  language understanding\BBCQ\
\newblock \url{https://arxiv.org/abs/2205.11487}.

\bibitem[\protect\BCAY{Sanh, Debut, Chaumond, \BBA\ Wolf}{Sanh
  et~al.}{2019}]{sanh2019distilbert}
Sanh, V., Debut, L., Chaumond, J., \BBA\ Wolf, T. \BBOP2019\BBCP.
\newblock \BBOQ DistilBERT, a distilled version of BERT: smaller, faster,
  cheaper and lighter\BBCQ\
\newblock \url{https://arxiv.org/abs/1910.01108}.

\bibitem[\protect\BCAY{Sanh, Webson, Raffel, Bach, \BBA\ Lintang~Sutawika}{Sanh
  et~al.}{2021}]{victor2022t0}
Sanh, V., Webson, A., Raffel, C., Bach, S.~H., \BBA\ Lintang~Sutawika, e.~a.
  \BBOP2021\BBCP.
\newblock \BBOQ Multitask Prompted Training Enables Zero-Shot Task
  Generalization\BBCQ\
\newblock \url{https://arxiv.org/abs/2110.08207}.

\bibitem[\protect\BCAY{Shoeybi, Patwary, Puri, LeGresley, Casper, \BBA\
  Catanzaro}{Shoeybi et~al.}{2019}]{shoeybi2019megatron}
Shoeybi, M., Patwary, M., Puri, R., LeGresley, P., Casper, J., \BBA\ Catanzaro,
  B. \BBOP2019\BBCP.
\newblock \BBOQ Megatron-lm: Training multi-billion parameter language models
  using model parallelism\BBCQ\
\newblock \url{https://arxiv.org/abs/1909.08053}.

\bibitem[\protect\BCAY{Shuster, Komeili, Adolphs, Roller, Szlam, \BBA\
  Weston}{Shuster et~al.}{2022a}]{shuster2022language}
Shuster, K., Komeili, M., Adolphs, L., Roller, S., Szlam, A., \BBA\ Weston, J.
  \BBOP2022a\BBCP.
\newblock \BBOQ Language models that seek for knowledge: Modular search \&
  generation for dialogue and prompt completion\BBCQ\
\newblock \url{https://arxiv.org/abs/2203.13224}.

\bibitem[\protect\BCAY{Shuster, Xu, Komeili, Ju, Smith, Roller, Ung, Chen,
  Arora, Lane, et~al.}{Shuster et~al.}{2022b}]{shuster2022blenderbot}
Shuster, K., Xu, J., Komeili, M., Ju, D., Smith, E.~M., Roller, S., Ung, M.,
  Chen, M., Arora, K., Lane, J., et~al. \BBOP2022b\BBCP.
\newblock \BBOQ Blenderbot 3: a deployed conversational agent that continually
  learns to responsibly engage\BBCQ\
\newblock \url{https://arxiv.org/abs/2208.03188}.

\bibitem[\protect\BCAY{Smith, Patwary, Norick, LeGresley, Rajbhandari, Casper,
  Liu, Prabhumoye, Zerveas, Korthikanti, et~al.}{Smith
  et~al.}{2022}]{smith2022using}
Smith, S., Patwary, M., Norick, B., LeGresley, P., Rajbhandari, S., Casper, J.,
  Liu, Z., Prabhumoye, S., Zerveas, G., Korthikanti, V., et~al. \BBOP2022\BBCP.
\newblock \BBOQ Using deepspeed and megatron to train megatron-turing nlg 530b,
  a large-scale generative language model\BBCQ\
\newblock \url{https://arxiv.org/abs/2201.11990}.

\bibitem[\protect\BCAY{Soltan, Ananthakrishnan, FitzGerald, Gupta, Hamza, Khan,
  Peris, Rawls, Rosenbaum, Rumshisky, Prakash, Sridhar, Triefenbach, Verma,
  Tur, \BBA\ Natarajan}{Soltan et~al.}{2022}]{soltan2022alexatm}
Soltan, S., Ananthakrishnan, S., FitzGerald, J., Gupta, R., Hamza, W., Khan,
  H., Peris, C., Rawls, S., Rosenbaum, A., Rumshisky, A., Prakash, C.~S.,
  Sridhar, M., Triefenbach, F., Verma, A., Tur, G., \BBA\ Natarajan, P.
  \BBOP2022\BBCP.
\newblock \BBOQ AlexaTM 20B: Few-Shot Learning Using a Large-Scale Multilingual
  Seq2Seq Model\BBCQ\
\newblock \url{https://arxiv.org/abs/2208.01448}.

\bibitem[\protect\BCAY{Stokel-Walker \BBA\ Noorden}{Stokel-Walker \BBA\
  Noorden}{2023}]{stokelwalker2023nature}
Stokel-Walker, C.\BBACOMMA\  \BBA\ Noorden, R.~V. \BBOP2023\BBCP.
\newblock \BBOQ What ChatGPT and generative AI mean for science\BBCQ\
\newblock \url{https://www.nature.com/articles/d41586-023-00340-6}.

\bibitem[\protect\BCAY{Su, Shi, Kasai, Wang, Hu, Ostendorf, Yih, Smith,
  Zettlemoyer, \BBA\ Yu}{Su et~al.}{2022}]{su2022instructor}
Su, H.~S., Shi, W.~S., Kasai, J., Wang, Y., Hu, Y., Ostendorf, M., Yih, W.-t.,
  Smith, N.~A., Zettlemoyer, L., \BBA\ Yu, T. \BBOP2022\BBCP.
\newblock \BBOQ One Embedder, Any Task: Instruction-Finetuned Text
  Embeddings\BBCQ\
\newblock \url{https://arxiv.org/abs/2212.09741}.

\bibitem[\protect\BCAY{Taori, Gulrajani, Zhang, Dubois, Li, Guestrin, Liang,
  \BBA\ Hashimoto}{Taori et~al.}{2023}]{alpaca}
Taori, R., Gulrajani, I., Zhang, T., Dubois, Y., Li, X., Guestrin, C., Liang,
  P., \BBA\ Hashimoto, T.~B. \BBOP2023\BBCP.
\newblock \BBOQ Stanford Alpaca: An Instruction-following LLaMA model\BBCQ\
\newblock \url{https://github.com/tatsu-lab/stanford_alpaca}.

\bibitem[\protect\BCAY{Tay, Dehghani, Tran, Garcia, Bahri, Schuster, Zheng,
  Houlsby, \BBA\ Metzler}{Tay et~al.}{2022}]{yi2022ul2}
Tay, Y., Dehghani, M., Tran, V.~Q., Garcia, X., Bahri, D., Schuster, T., Zheng,
  H.~S., Houlsby, N., \BBA\ Metzler, D. \BBOP2022\BBCP.
\newblock \BBOQ Unifying Language Learning Paradigms\BBCQ\
\newblock \url{https://arxiv.org/abs/2205.05131}.

\bibitem[\protect\BCAY{Taylor, Kardas, Cucurull, Scialom, Hartshorn, Saravia,
  Poulton, Kerkez, \BBA\ Stojnic}{Taylor et~al.}{2022}]{taylor2022galactica}
Taylor, R., Kardas, M., Cucurull, G., Scialom, T., Hartshorn, A., Saravia, E.,
  Poulton, A., Kerkez, V., \BBA\ Stojnic, R. \BBOP2022\BBCP.
\newblock \BBOQ GALACTICA: A Large Language Model for Science\BBCQ\
\newblock \url{https://arxiv.org/abs/2211.09085}.

\bibitem[\protect\BCAY{Thoppilan, De~Freitas, Hall, Shazeer, Kulshreshtha,
  Cheng, Jin, Bos, Baker, Du, et~al.}{Thoppilan
  et~al.}{2022}]{thoppilan2022lamda}
Thoppilan, R., De~Freitas, D., Hall, J., Shazeer, N., Kulshreshtha, A., Cheng,
  H.-T., Jin, A., Bos, T., Baker, L., Du, Y., et~al. \BBOP2022\BBCP.
\newblock \BBOQ Lamda: Language models for dialog applications\BBCQ\
\newblock \url{https://arxiv.org/abs/2201.08239}.

\bibitem[\protect\BCAY{Touvron, Lavril, Izacard, Martinet, Lachaux, Lacroix,
  Rozière, Goyal, Hambro, Azhar, Rodriguez, Joulin, Grave, \BBA\
  Lample}{Touvron et~al.}{2023}]{touvron2023llama}
Touvron, H., Lavril, T., Izacard, G., Martinet, X., Lachaux, M.-A., Lacroix,
  T., Rozière, B., Goyal, N., Hambro, E., Azhar, F., Rodriguez, A., Joulin,
  A., Grave, E., \BBA\ Lample, G. \BBOP2023\BBCP.
\newblock \BBOQ LLaMA: Open and Efficient Foundation Language Models\BBCQ\
\newblock \url{https://arxiv.org/abs/2302.13971}.

\bibitem[\protect\BCAY{Vaswani, Shazeer, Parmar, Uszkoreit, Jones, Gomez,
  Kaiser, \BBA\ Polosukhin}{Vaswani et~al.}{2017}]{vaswani2017attention}
Vaswani, A., Shazeer, N., Parmar, N., Uszkoreit, J., Jones, L., Gomez, A.~N.,
  Kaiser, {\L}., \BBA\ Polosukhin, I. \BBOP2017\BBCP.
\newblock \BBOQ Attention is all you need\BBCQ\
\newblock {\Bem Advances in Neural Information Processing Systems}, {\Bem 30}.

\bibitem[\protect\BCAY{Wang \BBA\ Komatsuzaki}{Wang \BBA\
  Komatsuzaki}{2021}]{gpt-j}
Wang, B.\BBACOMMA\  \BBA\ Komatsuzaki, A. \BBOP2021\BBCP.
\newblock \BBOQ {GPT-J-6B: A 6 Billion Parameter Autoregressive Language
  Model}\BBCQ\
\newblock \url{https://github.com/kingoflolz/mesh-transformer-jax}.

\bibitem[\protect\BCAY{Wang, Yang, Huang, Jiao, Yang, Jiang, Majumder, \BBA\
  Wei}{Wang et~al.}{2022}]{wang2022e5}
Wang, L., Yang, N., Huang, X., Jiao, B., Yang, L., Jiang, D., Majumder, R.,
  \BBA\ Wei, F. \BBOP2022\BBCP.
\newblock \BBOQ Text Embeddings by Weakly-Supervised Contrastive
  Pre-training\BBCQ\
\newblock \url{https://arxiv.org/abs/2212.03533}.

\bibitem[\protect\BCAY{Yang, Zhang, Song, Hong, Xu, Zhao, Shao, Zhang, Cui,
  \BBA\ Yang}{Yang et~al.}{2022}]{yang2022diffusion}
Yang, L., Zhang, Z., Song, Y., Hong, S., Xu, R., Zhao, Y., Shao, Y., Zhang, W.,
  Cui, B., \BBA\ Yang, M.-H. \BBOP2022\BBCP.
\newblock \BBOQ Diffusion models: A comprehensive survey of methods and
  applications\BBCQ\
\newblock \url{https://arxiv.org/abs/2209.00796}.

\bibitem[\protect\BCAY{Yang, Dai, Yang, Carbonell, Salakhutdinov, \BBA\
  Le}{Yang et~al.}{2019}]{yang2019xlnet}
Yang, Z., Dai, Z., Yang, Y., Carbonell, J., Salakhutdinov, R.~R., \BBA\ Le,
  Q.~V. \BBOP2019\BBCP.
\newblock \BBOQ Xlnet: Generalized autoregressive pretraining for language
  understanding\BBCQ\
\newblock {\Bem Advances in Neural Information Processing Systems}, {\Bem 32}.

\bibitem[\protect\BCAY{Zaheer, Guruganesh, Dubey, Ainslie, Alberti, Ontanon,
  Pham, Ravula, Wang, Yang, et~al.}{Zaheer et~al.}{2020}]{zaheer2020big}
Zaheer, M., Guruganesh, G., Dubey, K.~A., Ainslie, J., Alberti, C., Ontanon,
  S., Pham, P., Ravula, A., Wang, Q., Yang, L., et~al. \BBOP2020\BBCP.
\newblock \BBOQ Big bird: Transformers for longer sequences\BBCQ\
\newblock {\Bem Advances in Neural Information Processing Systems}, {\Bem 33},
  17283--17297.

\bibitem[\protect\BCAY{Zhang, Zhao, Saleh, \BBA\ Liu}{Zhang
  et~al.}{2020}]{zhang2020pegasus}
Zhang, J., Zhao, Y., Saleh, M., \BBA\ Liu, P. \BBOP2020\BBCP.
\newblock \BBOQ Pegasus: Pre-training with extracted gap-sentences for
  abstractive summarization\BBCQ\
\newblock In {\Bem International Conference on Machine Learning}, \BPGS\
  11328--11339. PMLR.

\bibitem[\protect\BCAY{Zhang, Roller, Goyal, Artetxe, Chen, Chen, Dewan, Diab,
  Li, Lin, et~al.}{Zhang et~al.}{2022}]{zhang2022opt}
Zhang, S., Roller, S., Goyal, N., Artetxe, M., Chen, M., Chen, S., Dewan, C.,
  Diab, M., Li, X., Lin, X.~V., et~al. \BBOP2022\BBCP.
\newblock \BBOQ Opt: Open pre-trained transformer language models\BBCQ\
\newblock \url{https://arxiv.org/abs/2205.01068}.

\bibitem[\protect\BCAY{Zhang, Sun, Galley, Chen, Brockett, Gao, Gao, Liu, \BBA\
  Dolan}{Zhang et~al.}{2019a}]{zhang2019dialogpt}
Zhang, Y., Sun, S., Galley, M., Chen, Y.-C., Brockett, C., Gao, X., Gao, J.,
  Liu, J., \BBA\ Dolan, B. \BBOP2019a\BBCP.
\newblock \BBOQ Dialogpt: Large-scale generative pre-training for
  conversational response generation\BBCQ\
\newblock \url{https://arxiv.org/abs/1911.00536}.

\bibitem[\protect\BCAY{Zhang, Han, Liu, Jiang, Sun, \BBA\ Liu}{Zhang
  et~al.}{2019b}]{zhang2019ernie}
Zhang, Z., Han, X., Liu, Z., Jiang, X., Sun, M., \BBA\ Liu, Q. \BBOP2019b\BBCP.
\newblock \BBOQ ERNIE: Enhanced language representation with informative
  entities\BBCQ\
\newblock \url{https://arxiv.org/abs/1905.07129}.

\end{thebibliography}
\bibliographystyle{theapa} 

\end{document}